\documentclass[review,3p]{elsarticle}
\usepackage{lscape}
\usepackage[utf8]{inputenc} 
\usepackage[T1]{fontenc}    
\usepackage{hyperref}       
\usepackage{url}            
\usepackage{booktabs}       
\usepackage{amsfonts}       
\usepackage{nicefrac}       
\usepackage{microtype}      
\usepackage{xcolor}         
\usepackage{multicol}
\usepackage{multirow}
\usepackage{subcaption}
\definecolor{blue_jokin}{HTML}{00F9DE}

\usepackage{algorithm}
\usepackage{algorithmic}
\usepackage{wrapfig}
\usepackage{graphicx}
\usepackage{mathtools} 
\usepackage{amssymb}
\usepackage{amsmath}
\usepackage{amsthm} 
\usepackage{color}
\usepackage{bm}

\newtheorem{theorem}{Theorem}

\newtheorem{definition}{Definition}
\usepackage{tcolorbox}

\usepackage{natbib}

\usepackage{color}

\newcommand{\bme}{\bm{e}}

\newcommand{\bmx}{\bm{x}}

\newcommand{\rmc}{\mathrm{c}}

\newcommand{\rme}{\mathrm{e}}

\newcommand{\rmi}{\mathrm{i}}

\newcommand{\sfA}{\mathsf{A}}
\newcommand{\sfB}{\mathsf{B}}

\newcommand{\sfH}{\mathsf{H}}

\newcommand{\sfK}{\mathsf{K}}

\newcommand{\sfQ}{\mathsf{Q}}

\newcommand{\sfU}{\mathsf{U}}
\newcommand{\sfV}{\mathsf{V}}
\newcommand{\sfW}{\mathsf{W}}
\newcommand{\sfX}{\mathsf{X}}

\newcommand{\calD}{\mathcal{D}}

\newcommand{\calM}{\mathcal{M}}

\journal{Knowledge-Based Systems}

\begin{document}

\begin{frontmatter}

\title{Diagnostic Spatio-temporal Transformer with Faithful Encoding}

\author[inst1,inst2]{Jokin Labaien}

\affiliation[inst1]{organization={Ikerlan Technology Research Centre, Basque Research and Technology Alliance (BRTA)},
            addressline={Pº J.M. Arizmediarrieta, 2}, 
            city={Arrasate/Mondrag\'o},
            postcode={20500}, 
            state={Gipuzkoa},
            country={Spain}}

\author[inst3]{Tsuyoshi Id\'e}
\author[inst3]{Pin-Yu Chen}
\author[inst2]{Ekhi Zugasti}
\author[inst1]{Xabier De Carlos}

\affiliation[inst2]{organization={Mondragon University},
            addressline={ Goiru Kalea, 2}, 
            city={Arrasate/Mondrag\'o},
            postcode={20500}, 
            state={Gipuzkoa},
            country={Spain}}
            
\affiliation[inst3]{organization={IBM Research},
            addressline={Kitchawan Rd}, 
            city={Yorktown Heights},
            postcode={10598}, 
            state={New York},
            country={United States}}

\begin{abstract} \label{abstrac}
This paper addresses the task of anomaly diagnosis when the underlying data generation process has a complex spatio-temporal (ST) dependency. The key technical challenge is to extract actionable insights from the dependency tensor characterizing high-order interactions among temporal and spatial indices. We formalize the problem as supervised dependency discovery, where the ST dependency is learned as a side product of multivariate time-series classification. We show that temporal positional encoding used in existing ST transformer works has a serious limitation 
in capturing higher frequencies (short time scales). We propose a new positional encoding with a theoretical guarantee, based on discrete Fourier transform. We also propose a new ST dependency discovery framework, which can provide readily consumable diagnostic information in both spatial and temporal directions. Finally, we demonstrate the utility of the proposed model, DFStrans (Diagnostic Fourier-based Spatio-temporal Transformer), in a real industrial application of building elevator control. 
\end{abstract}



\begin{keyword}
Anomaly Diagnosis \sep Spatio-Temporal Dependencies \sep Multi-sensor Data
\end{keyword}

\end{frontmatter}


\section{Introduction} \label{sec:intro}



With the recent advances in sensing technologies, anomaly detection has become critical in Industry 4.0 applications. \textit{Anomaly diagnosis}, which aims at getting explanations of detected anomalies in some way, is one of the interesting topics that modern highly sensorized industrial systems have brought into the machine learning community. 
Anomaly diagnosis from noisy multivariate sensor signals is known to be notoriously challenging. One of the major reasons is that we do not know upfront what spatial and temporal scales are relevant to anomalies we are interested in. Some failures may be characterized by high-frequency jitters, while others may be due to relatively long-term shifts of certain signals. Also, some failures may be detected as an outlier of a single sensor, while other failures may involve multiple sensors. The key technical challenge in anomaly diagnosis is to automatically learn the higher-order interactions between the temporal and spatial indices inherent in the multi-sensor data \cite{li2021stacking}. 

For flexibly capturing various temporal scales and dependencies,  Transformer networks look like a promising approach. Recently, \textit{spatio-temporal} (ST) extensions of the Transformer model have attracted attention in a variety of tasks such as human motion tracking and traffic analysis. Unlike conventional convolutional networks~\cite{tran2015learning}, Transformer networks have the potential to capture complex ST dependencies in a highly interpretable fashion, making them an attractive choice for anomaly diagnosis. There have been mainly two approaches in ST transformers in the literature: One is the \textit{twin} approach, which builds two transformer networks for the spatial and temporal dimensions~\cite{aksan2020attention,liu2021gated,yan2021learning}. The other is the \textit{hybrid} approach, which typically combines a graph convolutional network (GCN) with a temporal transformer~\cite{cai2020traffic,yu2020spatio,grigsby2021long}.  

Although those works report promising results in their targeted applications, there are three major limitations in our context. \textit{First}, the hybrid transformer approach requires prior knowledge of the spatial dependency, typically in the form of physical distance, which is not available in most industrial anomaly diagnosis settings. \textit{Second}, the twin approach uses essentially the same sinusoidal positional encoding as proposed in~\cite{vaswani2017attention} without clear justification. \textit{Third}, their main focus is temporal forecasting or its variant, and hence, they do not provide readily consumable information for anomaly diagnosis.

\begin{figure*}[t]
    \centering
    \includegraphics[width=\textwidth]{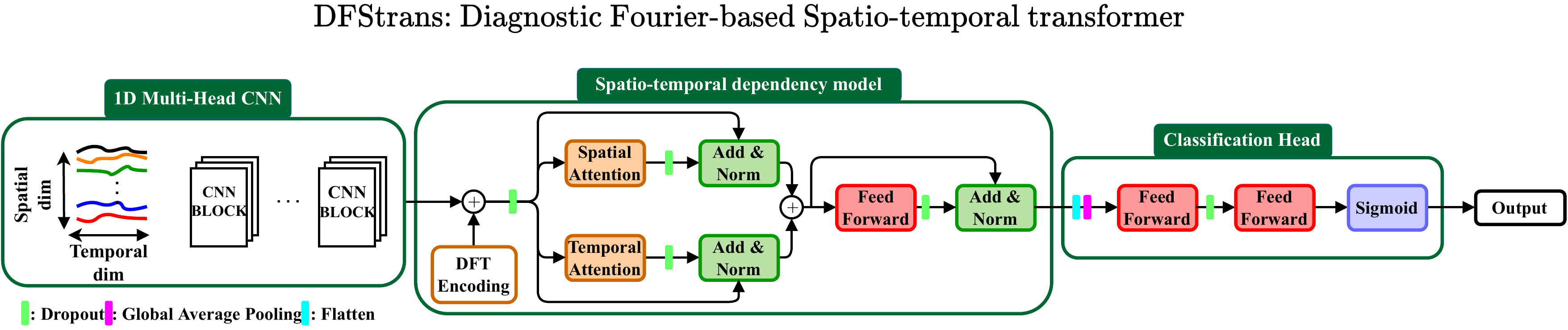}
    \caption{Architecture of the proposed DFStrans.}
    \label{fig:st_transformer}
\end{figure*}

In this paper, we formalize the anomaly diagnosis problem as supervised dependency discovery, where ST dependencies are learned as a side product of a multivariate time-series classification task (see Sec.~\ref{sec:problem_setting} for the detail). Our framework features a novel temporal encoding approach by combining a new positional encoding algorithm with a 1D multi-head CNN. We mathematically show that conventional sinusoidal encoding has a serious limitation in capturing short-scale variability. As outlined in Fig.~\ref{fig:st_transformer}, an embedded time-series episode enriched by the positional encoding is passed to a ST dependency model. Thanks to a proposed ST factorization model, our framework can provide local and global diagnostic scores, which can be readily consumed in real-world anomaly diagnosis tasks. Unlike the existing hybrid ST transformer approaches, it can also learn spatial dependency without prior knowledge.

In summary, our contributions are the proposal of (1) a new positional encoding algorithm called the DFT (discrete Fourier transform) encoding that has a faithfulness guarantee and (2) an ST diagnostic approach based on the proposed ST dependency discovery model. 

\section{Related work}

\paragraph{Multivariate time-series anomaly detection} Multivariate time series anomaly detection is essential for identifying unusual patterns that may indicate a problem. Deep learning models are frequently used in the literature for this purpose. Some of these models are deterministic, such as the LSTM-based Encoder-Decoder scheme proposed in~\cite{malhotra2016lstm} which learns to reconstruct normal behavior and uses reconstruction error to detect anomalies, or the hybrid approach combining Multi-Head CNNs and recurrent networks for anomaly detection in multi-sensor industrial data described in~\cite{Canizo2019}. Other models are stochastic, such as those described in~\cite{su2019robust} and~\cite{an2015variational}. In~\cite{su2019robust} (\textit{OmniAnomaly}), the authors first use an autoregressive model to model the temporal dependencies of each variable in the time series data, then utilize a variational autoencoder (VAE) to reconstruct the data points and identify discrepancies between the original data and the reconstructed data. VAEs are a common choice for modeling normal behavior in anomaly detection~\cite{an2015variational}. Other generative methods, such as Generative Adversarial Networks (GANs), have also been utilized for the purpose of detecting anomalies~\cite{liu2022time}. Recently, Zhang et al.~\cite{zhang2023stad} proposed a novel self-training GAN, called STAD-GAN, which has been shown to be effective in the presence of noisy training data. In this work, anomaly detection is achieved by using the generator to capture the normal data distribution and the discriminator to amplify the reconstruction error for abnormal data. For a general overview and more recent works, see~\cite{blazquez2021review,pang2021deep}.

\paragraph{Transformers for anomaly detection} Anomaly detection via transformers is an emerging field nowadays. Meng et al. \cite{meng2019spacecraft} propose to use an encoder-decoder transformer architecture to capture anomalies in multivariate time-series data unsupervisedly. In this approach, they use a masking strategy to reconstruct time-series, and then they use a non-dynamic threshold \cite{malhotra2015long} to detect anomalies via reconstruction error. For anomaly detection in system logs, Huang et al. proposed \textit{HitAnomaly}  \cite{huang2020hitanomaly}, a hierarchical transformer that captures semantic information from both log templates and parameter values that are merged to make a classification based on attention mechanisms. Chen et al. \cite{chen2021learning} propose \textit{GTA}, a method that uses a graph structure to learn intra-sensor relationships from multi-sensor data, modeling temporality with transformers and detecting anomalies using a fixed threshold over the reconstruction error. Recently, Xu et al.\cite{xu2021anomaly} proposed \textit{AnomalyTransformer}, a transformer specially designed for anomaly detection that introduces an attention mechanism employing a min-max strategy to amplify the difference between normal and abnormal time points. In \cite{tuli2022tranad} \cite{tuli2022tranad}, a transformer-based model, TranAD, is proposed for efficiently detecting and diagnosing anomalies in multivariate time series data. TranAD uses attention-based sequence encoders to perform inference with knowledge of the broader temporal trends in the data, and employs focus score-based self-conditioning and adversarial training to facilitate robust multi-modal feature extraction and enhance stability. The authors also employ model-agnostic meta learning (MAML) to train the model using limited data. In~\cite{doshi2022reward}, the authors propose an encoder-decoder transformer architecture for detecting anomalies, but in this case, they use the sparse self-attention proposed in Informer~\cite{zhou2021informer}, which allows focusing on the most important queries rather than all of them. The authors also propose a new way of evaluating anomaly detection models. Hybrid models combining Transformer models with 1D convolutions has recently proven to be successful in anomaly detection, as pointed in~\cite{kim2023time}, achieving great results with an architecture that includes multiple Transformer encoder layers and a decoder with 1D convolution layers.

\paragraph{Spatio-temporal transformers} Spatio-temporal transformers are a type of transformer that can capture spatio-temporal relationships within data and have been applied to tasks such as visual tracking \cite{yan2021learning} and human motion prediction \cite{aksan2020attention, aksan2020spatio}. In order to capture spatial dependencies in multivariate time series data, several transformer-based approaches have been proposed that utilize graph neural networks \cite{yu2020spatio, cai2020traffic}. The \textit{NAST} model \cite{chen2021nast} applies a spatio-temporal attention mechanism to capture spatial dependencies in parallel and then combines them for time-series forecasting. The "3D Spatio-Temporal Transformer" \cite{aksan2020attention} uses decoupled 3D temporal and spatial attention blocks to learn intra- and inter-articular dependencies over time. Liu et al. \cite{liu2021gated} propose a combination of transformer networks with a gating mechanism for time-series classification. These models are often used to identify relationships between variables in multivariate time series data. In time-series forecasting, Huang et al.~\cite{huang2022spatial} proposes a spatio-temporal transformer that uses 1D convolutions. To capture complex spatio-temporal dependencies, the paper proposes the use of local-range convolutional attention and group-range convolutional attention.

\paragraph{Encoding strategies in Transformers} Transformers use positional encoding to introduce information about the position of items in a sequence. Vanilla positional encodings use different frequency sine waves to model this, but as we will show later, this can be problematic in scenarios where small changes in time matter. Currently, some works propose the use of discrete Fourier transform (DFT) in positional encodings as an alternative. One typical approach is to use DFT or wavelet transform to compress input time-series~\cite{mao2019learning,pan2021spatiotemporal,zhou2022fedformer}, which is not appropriate for anomaly diagnosis, where interpretability matters. \cite{kazemi2019time2vec} is close to our work in its spirit, but differs in its motivation (capturing periodic components) and the lack of faithfulness guarantee. \citet{buchholz2022fourier} re-formalized the transformer in the Fourier space in an image-specific way, which makes it inapplicable to our setting. \citet{aksan2021spatio} proposed a spatio-temporal transformer but with the vanilla positional encoding for the temporal dimension. 



\section{Problem setting} \label{sec:problem_setting}
This section states the problem setting and explains the dependency discovery problem.

\paragraph{\textbf{Training data}}

  \begin{figure}[!ht]
    \centering
    \includegraphics[width=0.6\textwidth]{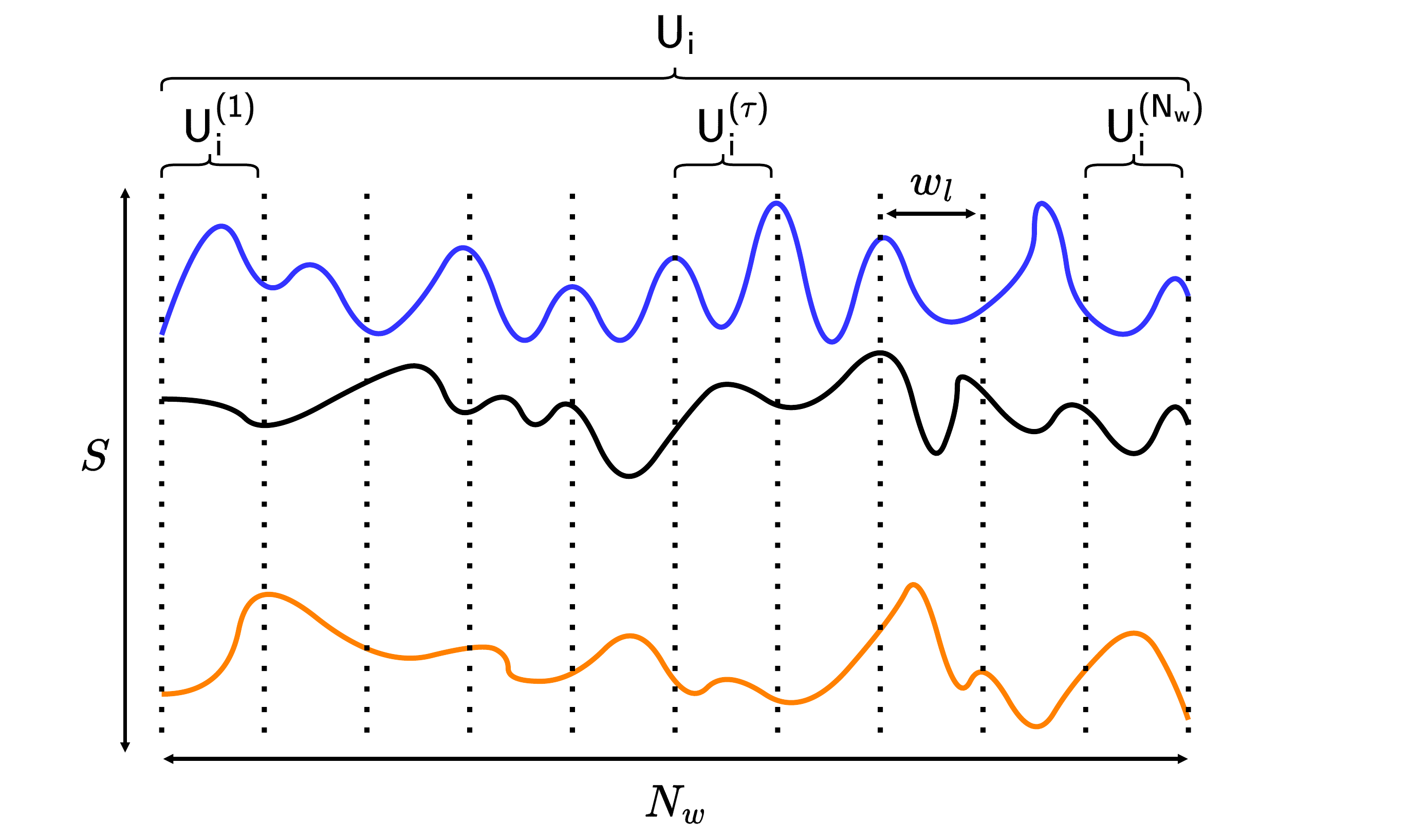}
    \caption{Illustration of the notation used. One episode of $S$-variate time-series is split into $N_w$ consecutive segments of size $w_l$. A binary label $y_i$ is associated with the time-series episode $\sfU_i$, where $y_i=1$ means that the episode $\sfU_i$ is anomalous.}
      \label{fig:notation}
  \end{figure}

We are given $N$ sets of $S$-variate time-series and a binary label as 
$\calD_{\mathrm{train}} \triangleq\{(\sfU_i,y_i) \mid i = 1,\ldots,N\}$. 
Each of the $S$ time-series typically corresponds to measurements of one physical sensor. The binary label $y_i$ is 1 if $\sfU_i$ is anomalous and 0 otherwise. Although anomalous samples are generally hard to obtain in many real-world anomaly \textit{detection} settings, we assume that reasonably many positive samples have been collected for the purpose of anomaly \textit{diagnosis}. Each time-series episode is divided into $N_w$ consecutive segments using a sliding window of size $w_l$ as $\sfU_i = [\sfU_i^{(1)}, \ldots, \sfU_i^{(N_w)}]$, where $\sfU_i^{(\tau)} \in \mathbb{R}^{S\times w_l}$ is the $\tau$-th time-segment. An example of a time-series episode is an ascend or descend session of an elevator. Note that both temporal and spatial dependencies matter within one episode, while different episodes are considered statistically independent.  Unlike human motion tracking and traffic monitoring~\cite{aksan2020spatio,cai2020traffic} we do not assume prior knowledge on the similarity in the spatial dimension (i.e., among sensors). See Figure \ref{fig:notation} for a detailed overview of the notation used.



\paragraph{\textbf{Supervised dependency discovery problem}} 
From a practical perspective, our goal is to obtain actionable insights into anomalous samples by providing explanations of how they are anomalous. Specifically, major questions we wish to answer include (1) at a moment that an unusual situation is observed in one or a few variable(s), how the other variables are related, and (2) in a sensor showing an unusual behavior, how the unusual behavior might have been triggered by past events. 


As a machine learning task, this can be done by solving the \textit{supervised dependency discovery} problem. Specifically, we train a binary classifier using a prediction model that incorporates a learnable spatio-temporal (ST) dependency model (explained in Sec.~\ref{subsec:dependency_model}) as part of the data generative process. If the classification accuracy is high enough, the learned dependency can be used as a proxy of real dependency structure. Although our objective function is simply the binary cross-entropy (BCE):
\begin{align}
    \ell[\calM] = -\sum_{i=1}^{N}  \{y_i \ln p( y_i=1 \mid \sfU_i) + (1- y_i)\ln[1- p( y_i=1 \mid \sfU_i)\},
\end{align} 
where $p( y_i=1 \mid \sfU_i)$ is the probability of $\sfU_i$ being anomalous (i.e. the output given by the sigmoid function) and $\calM$ symbolically represents the ST dependency model, our main motivation is to find $\calM$ through maximizing the classification performance.

\section{ Spatio-temporal dependency discovery framework}
The proposed anomaly diagnosis framework consists of three major components as shown in Fig.~\ref{fig:st_transformer}. This section first states the problem setting and explains the major components in detail.

\subsection{Overall model architecture}

The goal of automatically capturing informative ST patterns requires the capability of handling different temporal resolutions as well as learning ST dependencies. 

For the former, we employ a multi-head one-dimensional (1D) convolutional neural network (CNN), similar to the one used in~\cite{Canizo2019}, which consists of multi-head 1D convolution, MaxPooling, ReLU activation, and batch normalization, applied independently to each of the variables (see Appendix for the detail). This module defines a mapping from raw time-series segment to a representation matrix: $\sfU_i^{(\tau)} \to \sfX_i^{(\tau)} \in \mathbb{R}^{S\times M}$, where $M$ is the dimensionality of the representation vector. To highlight spatial-temporal aspects of the data, we mainly use $\{\bmx_i^{(\tau,s)}\}$ instead of $\sfX_i^{(\tau)}$, where $\bmx_i^{(\tau,s)}\ \in \mathbb{R}^M$ and $\tau,s$ run over $1,\ldots,N_w$ and $1,\ldots,S$, respectively. Unless confusion is likely, we omit the sample index $i$ from $\bmx_i^{(\tau,s)}$ hereafter.

As shown in Fig.~\ref{fig:st_transformer}, the output of the 1D CNN module is fed into the transformer module. The feature tensor $\{\bmx^{(\tau,s)}\}$ is transformed into an enriched version of the same size, as explained later. 

Finally, in the classification head, the output of the transformer is first flattened along the sensor dimension. A global average pooling~\cite{lin2013network} layer is then applied along the temporal dimension. The resulting vector is processed by a fully connected layer with ReLU activation and a dropout layer. The final output probability $p( y_i=1 \mid \sfU_i)$ is determined by a fully connected layer with a sigmoid activation function.


\subsection{Positional encoding}
\label{sec:pe_dft}

The goal of positional encoding is to reflect the information of the position of the items in a sequence. In the original formulation~\cite{vaswani2017attention}, this was done by simply adding an extra vector $\bme^{(\tau)}$ to the embedding vector of a word as $ 
    {\bmx}^{(\tau)} \leftarrow {\bmx}^{(\tau)}  + {\bme}^{(\tau)},
$ 
where ${\bmx}^{(\tau)}$ is the representation vector of the $\tau$-th word and $\bme^{(\tau)}$ is its corresponding positional encoding (no spatial index $s$ was considered so is omitted). They used the following form for $\bme^{(\tau)}$:
\begin{align}\label{eq:positional_encoding_vaswani512}
     \left(\sin (w_0\tau),
    \cos  (w_0\tau),
    \sin  (w_2\tau),
    \cos  (w_2\tau),\cdots, 
    \sin (w_{d-2}\tau),
    \cos (w_{d-2}\tau)
    \right)^\top
\end{align}
with $w_k =  \rho^{-\frac{k}{d}}$, $\rho=10\,000$, and $d=512$.

As discussed in Section~\ref{sec:Fourier_PE} in detail, one issue with the original positional encoding~\eqref{eq:positional_encoding_vaswani512} is that it has a strong low-pass filtering property. As a result, it puts too much emphasis on long-range correlation among items and tends to suppress short- and mid-range location differences. To address the issue, we propose to use what we call the \textbf{faithful-Encoding}:
\begin{align} \label{eq:positional_encoding_DFT}
  \!  \bme^{(\tau)}=\sqrt{\frac{2}{d}}\left(
\frac{1}{\sqrt{2}}, 
\cos(\omega_1\tau),
\sin(\omega_1\tau),\ldots,
\cos(\omega_K\tau),
\sin(\omega_K\tau),
\frac{\cos \pi \tau}{\sqrt{2}}
\right)^\top,
\end{align}
where $\omega_k \triangleq \frac{2\pi k}{d}$ and $K\triangleq \frac{d}{2}-1$. The derivation is given in Section~\ref{sec:Fourier_PE}. Although the faithful-Encoding seemingly looks similar to the original one~\eqref{eq:positional_encoding_vaswani512}, there is a fundamental difference. One can show that the particular form~\eqref{eq:positional_encoding_DFT} has a mathematical guarantee that it maintains the whole information of the location without introducing any bias.

\subsection{Learning spatio-temporal dependencies}\label{subsec:dependency_model}

Now that temporal ordering has been reflected by the faithful-Encoding, we next consider how to reflect ST dependencies in the representation vectors. Intuitively, this can be done by making highly dependent items have similar representation vectors. Mathematically, it is implemented by defining the transition probability between items in an input sequence.

\begin{figure}[!ht]
    \centering
    \includegraphics[width=0.4\textwidth]{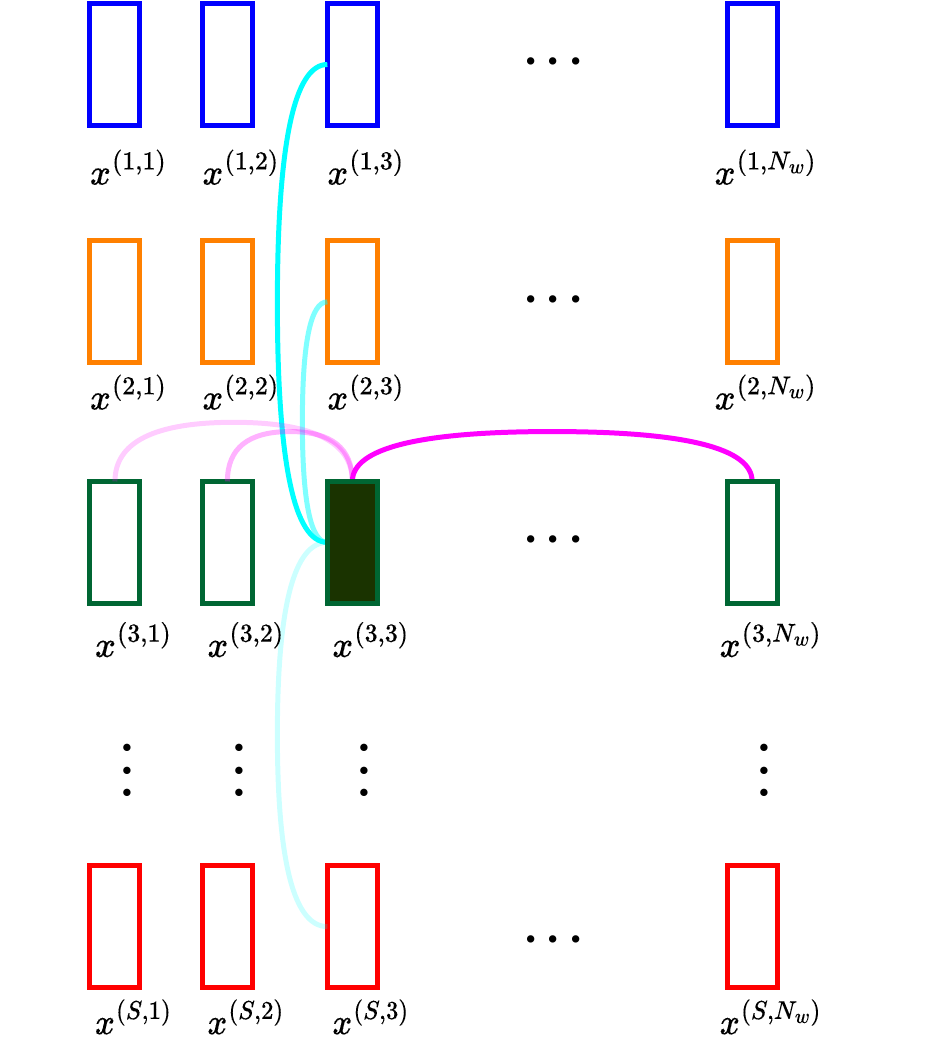}
    \caption{Each rectangular cell within a row (spatial dimension) and a column (temporal dimension) corresponds to an embedding. For a given sensor $s$ and time-segment $\tau$ (black rectangle), the \textcolor{magenta}{pink} lines denote temporal dependencies, while the \textcolor{blue_jokin}{blue} lines represent spatial dependencies. Notably, the intensity of the color in each cell reflects the strength of the corresponding dependency.}
    \label{fig:st_att}
\end{figure}

Our basic assumption is that there is a latent dependency structure behind  $\{\bmx^{(\tau,s)}\}$, which is described by the transition probability between two spatio-temporal points $(\tau,s)$ and $(\tau',s')$ as $p(\tau,s \mid \tau',s')$, as illustrated in Figure \ref{fig:st_att}.  Here we employ a factorized transition model 
\begin{align}
p(\tau,s \mid \tau',s') \approx p(\tau\mid \tau',s)p(s\mid s',\tau),
\end{align}
where $p(\cdot)$ is used as a symbolic notation representing probability distribution in general rather than a specific functional form. Each component is assumed to be log-quadratic:
\begin{align}
    \ln p(\tau\mid \tau',s) &= \rmc. + \frac{1}{\sqrt{M}} (\hat{\bmx}^{(\tau,s)})^\top \hat{\sfH} \text{ }\hat{\bmx}^{(\tau',s)},\\
    \ln p(s\mid s',\tau) &= \rmc. + \frac{1}{\sqrt{M}}(\overline{\bmx}^{(\tau,s)})^\top \overline{\sfH} \text{ } \overline{\bmx}^{(\tau,s')},
\end{align}

where $\rmc.$ symbolycally denotes a constant to meet the normalization condition of the probability distributions. Also, $\hat{\sfH},\overline{\sfH} \in \mathbb{R}^{M\times M}$ are fully learnable matrices that roughly correspond to the precision matrix of a Gaussian diffusion process in the feature space, and  $\hat{\bmx}^{(\tau,s)}, \overline{\bmx}^{(\tau,s)}$ are the temporal and spatial branches of the embeddings $\bmx^{(\tau,s)}$, respectively. The quadratic form can also be viewed as a generalized similarity between the pair of the feature vectors. In fact, if $\hat{\sfH},\overline{\sfH}$ are  identity matrices, the r.h.s.~is reduced to the well-known cosine similarity. The logarithm function guarantees the positivity of $p(\cdot)$. For more details on this, see Appendix.

So far, we have assumed that the representation vectors are given. Here, let us consider the opposite: If the transition probability is given, how can we find the representation vectors consistent with the transition probability? In the same spirit of graph neural networks, we can do this by making two-step Markovian transitions:
\begin{align}\label{eq:update-with-s-fixed}
    \hat{\bmx}^{(\tau,s)} &\leftarrow \sum_{\tau'} p(\tau\mid \tau',s)\hat{\sfW}_V \hat{\bmx}^{(\tau',s)} \quad \quad \mbox{for all $(\tau,s)$}, \\
    \label{eq:update-with-t-fixed}
    \overline{\bmx}^{(\tau,s)} &\leftarrow \sum_{s'} p(s\mid s', \tau)\overline{\sfW}_V\overline{\bmx}^{(\tau,s')} \quad \quad \mbox{for all $(\tau,s)$},
\end{align}
where $\hat{\sfW}_V,\overline{\sfW}_V \in \mathbb{R}^{M\times M}$ is a fully learnable parameter matrix to absorb indistinguishably due to scaling and rotation. As shown in Figure \ref{fig:st_transformer}, both $\hat{\bmx}^{(\tau,s)}$ and $\overline{\bmx}^{(\tau,s)}$ are added, obtaining an enriched representation $\bmx^{(\tau,s)}$. By this diffusion process, the higher the transition probability is between a pair of the representation vectors, the more similar they are.

\paragraph{\textbf{Relationship with Transformer}} \label{sec:a_b_mat}

Our ST dependency model can be viewed as a variant of the Transformer~\cite{vaswani2017attention}. To see this, consider the singular value decompositions (SVD) of the temporal and spatial branches $\hat{\sfH} = \hat{\sfW_Q} \hat{\sfW}_K^\top$ and  $\overline{\sfH} = \overline{\sfW}_Q\overline{\sfW}_K^\top$, where $\hat{\sfW}_Q$, $\hat{\sfW}_K$, $\overline{\sfW}_Q$ and $\overline{\sfW}_K$ are the left and right singular matrices with column vectors normalized to the square root of the singular values. Since any well-defined matrices have an SVD, without loss of generality, Eqs.~\eqref{eq:update-with-s-fixed}-\eqref{eq:update-with-t-fixed} are reduced to the well-known query-key-value mapping of the Transformer:
\begin{align}
    & \hat{\sfX}^{(s)} \leftarrow \mathrm{softmax}\left(\frac{1}{\sqrt{M}} \hat{\sfQ} \hat{\sfK}^\top\right)\hat{\sfV} \\ & \overline{\sfX}^{(\tau)} \leftarrow \mathrm{softmax}\left(\frac{1}{\sqrt{M}}\overline{\sfQ}  \overline{\sfK}^\top\right)\overline{\sfV}
\end{align}
where $\sfX^{(s)} \triangleq [\bmx^{(1,s)}, \ldots, \bmx^{(N_w,s)}]$,  $\hat{\sfQ}\triangleq \hat{\sfW}_Q\sfX^{(s)}, \hat{\sfK}\triangleq\hat{\sfW}_K\sfX^{(s)},\hat{\sfV} \triangleq \hat{\sfW}_V\sfX^{(s)}$  and $\sfX^{(\tau)} \triangleq [\bmx^{(\tau,1)}, \ldots, \bmx^{(\tau,S)}]$, $\overline{\sfQ}\triangleq \overline{\sfW}_Q\sfX^{(\tau)}, \ \overline{\sfK}\triangleq\overline{\sfW}_K\sfX^{(\tau)}, \overline{\sfV} \triangleq \overline{\sfW}_V\sfX^{(\tau)}$. After this, the enriched representations are added, normalized, and represented in a single representation vector:
\begin{align}
    \sfX  \leftarrow \operatorname{LayerNorm}(\sfX + \operatorname{Dropout}(\hat{\sfX})) + \operatorname{LayerNorm}(\sfX + \operatorname{Dropout}(\overline{\sfX}))
\end{align}
After the aggregation, the enriched representation $\sfX$ is normalized after passed to a 2-layer feedforward network with ReLU activation function and a dropout layer:
\begin{align}
    \sfX  \leftarrow \operatorname{LayerNorm}(\sfX + \operatorname{Dropout}(\operatorname{ReLU}\left((\sfX \sfW_1+b_1\right)\sfW_2+b_2)))
\end{align}
where $\sfW_1,\sfW_2 \in \mathbb{R}^{d \times d_{ff}}$ and $b_1, b_2 \in \mathbb{R}^{d_{ff}}$ are the weights and biases of the first and second feedforward layers, respectively. The dimension of the feedforward layer is set to $d_{ff} = 2048$. A single transformer layer is used in the experiments.

\subsection{Evaluation of attention.}

 As the model has been trained for anomaly detection, a high attention value should indicate that the model has given more importance to a given embedding in deciding whether it is anomalous. To evaluate if high attention means high impact in the model decision, we propose the \textit{Attention Trustworthiness Score} (\textit{AT-Score}). Th AT-Score evaluates the effectiveness of the attention mechanism in focusing correctly on when and where the anomalies have occurred by measuring how much the model prediction changes when the top-\textit{k} percent of the spatial and temporal attention weights are set to zero, and it is defined as
\begin{equation}
    \operatorname{AT}_k = \frac{1}{N_{test}} \sum_{i=1}^{N_{test}} \{\hat{y}(\sfU_i) -\hat{y}^k(\sfU_i)\}
\end{equation}
where $\hat{y}(\sfU_i)$ denotes the output of the model before sigmoid and $\hat{y}^k(\sfU_i) $ denotes the output  of the model before sigmoid and setting the top-\textit{k} percent of attention weights to zero.

\subsection{Diagnostic scores}

The learned transition probabilities allow a variety of diagnosis tasks. Let $\sfA$ and $\sfB$ be the temporal and spatial attention matrices. Based on these matrices we define some scores for temporal and spatial diagnosis.

For temporal diagnosis, we define four scores:
\begin{align}\label{eq:temporal_diag_scores}
& \sfA^{(s)}_{\tau', \tau}\triangleq p(\tau \mid \tau',s), \quad
a^{(s)}_\tau \triangleq \sum_{\tau'=1}^{N_w} \sfA^{(s)}_{\tau', \tau}, \quad
\sfA_{\tau',\tau}^{\mathrm{G}} \triangleq \frac{1}{S} \sum_{s=1}^{S} \sfA^{(s)}_{\tau', \tau}, \quad 
a^{\mathrm{G}}_\tau \triangleq \frac{1}{S}\sum_{s=1}^{S} a^{(s)}_\tau ,
\end{align}
where $a^{(s)}_\tau$ is the influence of a time point $\tau$ measured locally at the $s$-th sensor, $\sfA^G_{\tau',\tau}$ is the global influence of $\tau'$ on $\tau$, and $a^{\mathrm{G}}_\tau $ is the global influence of $\tau$ across all the variables.

For spatial (i.e.,~variable-variable) diagnosis, we define:
\begin{align}\label{eq:spatial_diag_scores}
& \sfB^{(\tau)}_{s',s}\triangleq p(s \mid s',\tau),\quad
b^{(\tau)}_s \triangleq \sum_{s'=1}^{S} \sfB^{(\tau)}_{s',s}, \quad 
\sfB_{s', s}^{\mathrm{G}} \triangleq \frac{1}{N_w} \sum_{\tau=1}^{N_w} \sfB^{(\tau)}_{s',s}a^{\mathrm{G}}_\tau, \quad
b^{\mathrm{G}}_s \triangleq \sum_{s'=1}^{S} \sfB^G_{s', s} , \quad
\end{align}
where $b^{(\tau)}_s $ is the influence of the variable $s$ at a time $\tau$, $\sfB^G_{s', s}$ is the global influence of the variable $s'$ on $s$, and $b^{\mathrm{G}}_s$ is the global influence of the variable~$s$.

\section{Fourier Analysis of Positional Encoding}
\label{sec:Fourier_PE}

In this section, we discuss how Fourier analysis is used to evaluate the goodness of positional encoding~(PE) and derive the faithful-Encoding~\eqref{eq:positional_encoding_DFT}.

\subsection{Original PE has a strong low-pass property}

Positional encoding in the present setting is the task of finding a $d$-dimensional representation vector of an item in the input sequence of length~$N_w$. In the vector view, the position of the $s$-th item in the input sequence is most straightfowardly represented by an $N_w$-dimensional ``one-hot'' vector, whose $s$-th entry is 1 and otherwise 0. Unfortunately, this is not an appropriate representation because the dimensionality is fixed to be $N_w$, and it does not have continuity over the elements at all, which makes numerical optimization challenging in stochastic gradient descent. 

The original PE in Eq.~\eqref{eq:positional_encoding_vaswani512} is designed to eliminate these limitations. Then, the question is how we can tell the goodness of its specific functional form. One approach suggested by the sinusoidal form is to use DFT. Consider DFT on the 1-dimensional (1D) lattice with $d$ lattice points ($d > N_w$). Any function defined on this lattice is represented as a linear combination of the sinusoidal function with the frequencies $\{\omega_k\}$. It is interesting to see how the frequencies $\{w_k\}$ in Eq.~\eqref{eq:positional_encoding_vaswani512} are distributed in the Fourier space.

\begin{figure}[bth]
\centering    
\includegraphics[trim={0cm 0cm 0cm 0cm},clip,width=10cm]{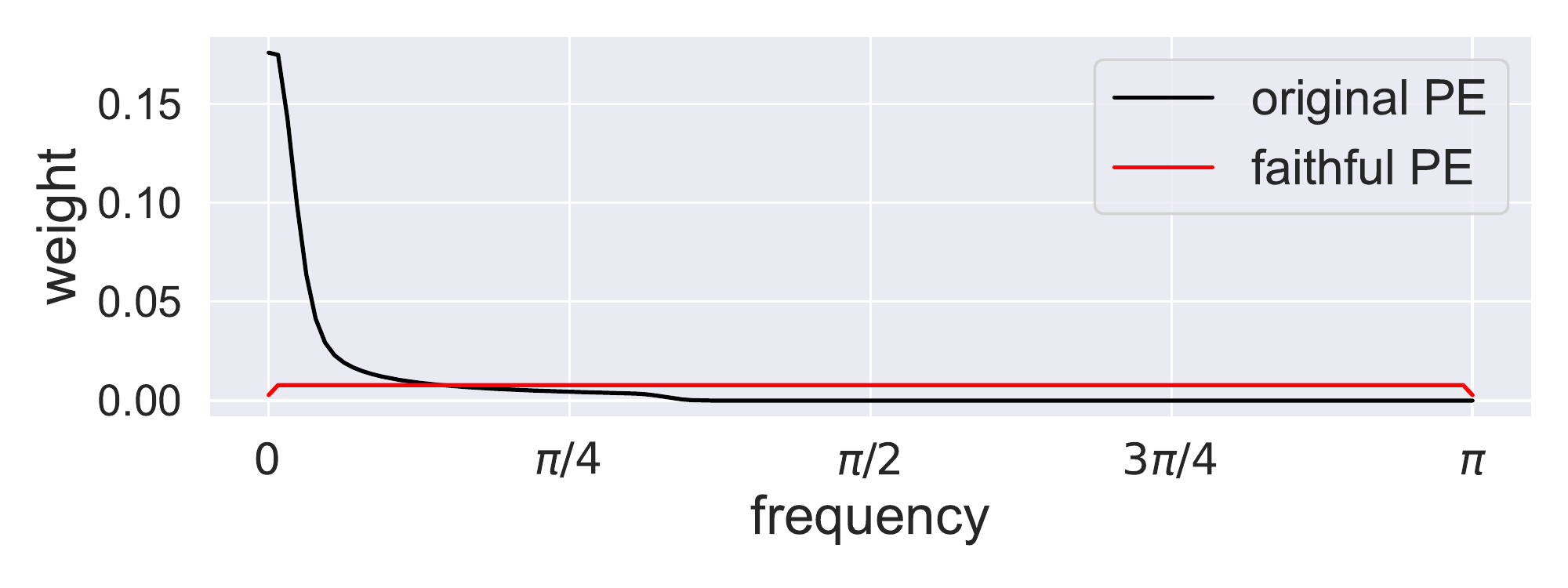}
    \caption{The distribution of the frequency $w_k = \rho^{-\frac{k}{d}}$ in Eq.~\eqref{eq:positional_encoding_vaswani512} over the Fourier bases (black line), where $\rho=10\,000,d=256$. Notice the contrast to that of the faithful-Encoding, which gives a uniform distribution (red line; See Section~\ref{sec:DFT_encoding}).}
    \label{fig:frequencyRatio}
    \vspace{-0.0cm}
\vspace{-0.0cm}
\end{figure}

The black line in Fig.~\ref{fig:frequencyRatio} shows the distribution of the frequencies of Eq.~\eqref{eq:positional_encoding_vaswani512}. The distribution is estimated with the kernel density estimation with the Gaussian kernel~\cite{Bishop}. Specifically, at each $\omega_k$, the weight is given by
\begin{align}
    g_k &= \sum_{l \in \{0,2,\ldots,d-2 \}}\frac{1}{R}\exp\left(-\frac{1}{2 \sigma^2}(\omega_k - w_l)^2\right),
\end{align}
where $R$ is a normalization constant to satisfy $\sum_k g_k=1$. We chose the bandwidth $\sigma=4\times \frac{2\pi}{d}$ with $d=256$. Due to the power function $\rho^{-\frac{k}{d}}$, the distribution is extremely skewed towards zero. 

This fact can be easily understood also by running a simple analysis as follows. The first and second smallest frequencies are $0, \frac{2\pi}{d}$, respectively. We can count the number of $w_k$s that fall into between them. Solving the equation
\begin{align}
    \frac{2\pi}{d} = \rho^{-\frac{l}{d}}
\end{align}
and assuming $\rho=10\,000$, we have $l = \frac{d}{4}\log_{10}\frac{d}{2\pi} \approx 103$ for $d=256$ and $l\approx 245$ for $d=512$. Hence, almost a half of the entries go to this lowest bin. 

This simple analysis, along with Fig.~\ref{fig:frequencyRatio}, demonstrates that the original PE has a strong bias to suppress mid and high frequencies. As a result, it tends to ignore mid- and short-range differences in location. This can be problematic in applications where short-range dependencies matter, such as time-series classification for physical sensor data.

\subsection{Original PE lacks faithfulness}

Another interesting question is what kind of function the skewed distribution $g_k$ represents. To answer this question, we perform an experiment described in Algorithm~\ref{algo:reference_reconstruction}, which is designed to understand what kind of distortion it may introduce to an assumed reference function. In our PE context, the reference function should be the position function (a.k.a.~one-hot vector) since the original PE~\eqref{eq:positional_encoding_vaswani512} was proposed to be a representation of the item at the location $s$.

\begin{algorithm}[tb]
\caption{Reference function reconstruction}
\label{algo:reference_reconstruction}
\begin{algorithmic}[1] 
\REQUIRE Reference function $f(t)$, DFT component weights $\{g_k\}$.
\STATE Find DFT coefficients of $f$ as $\{(a_k,b_k)\mid k=0,\ldots,K\}$.
\STATE $a_0 \leftarrow a_0 g_0$ and $b_0 \leftarrow b_0 g_{K+1}$
\FORALL{$k=1,\ldots,K$}
\STATE $a_k \leftarrow a_k g_k$ and $b_k \leftarrow b_k g_k$
\ENDFOR
\STATE Inverse-DFT from the modified coefficients.
\end{algorithmic}
\end{algorithm}

\begin{figure}[bth]
\centering    
\includegraphics[trim={0cm 0cm 0cm 0cm},clip,width=10cm]{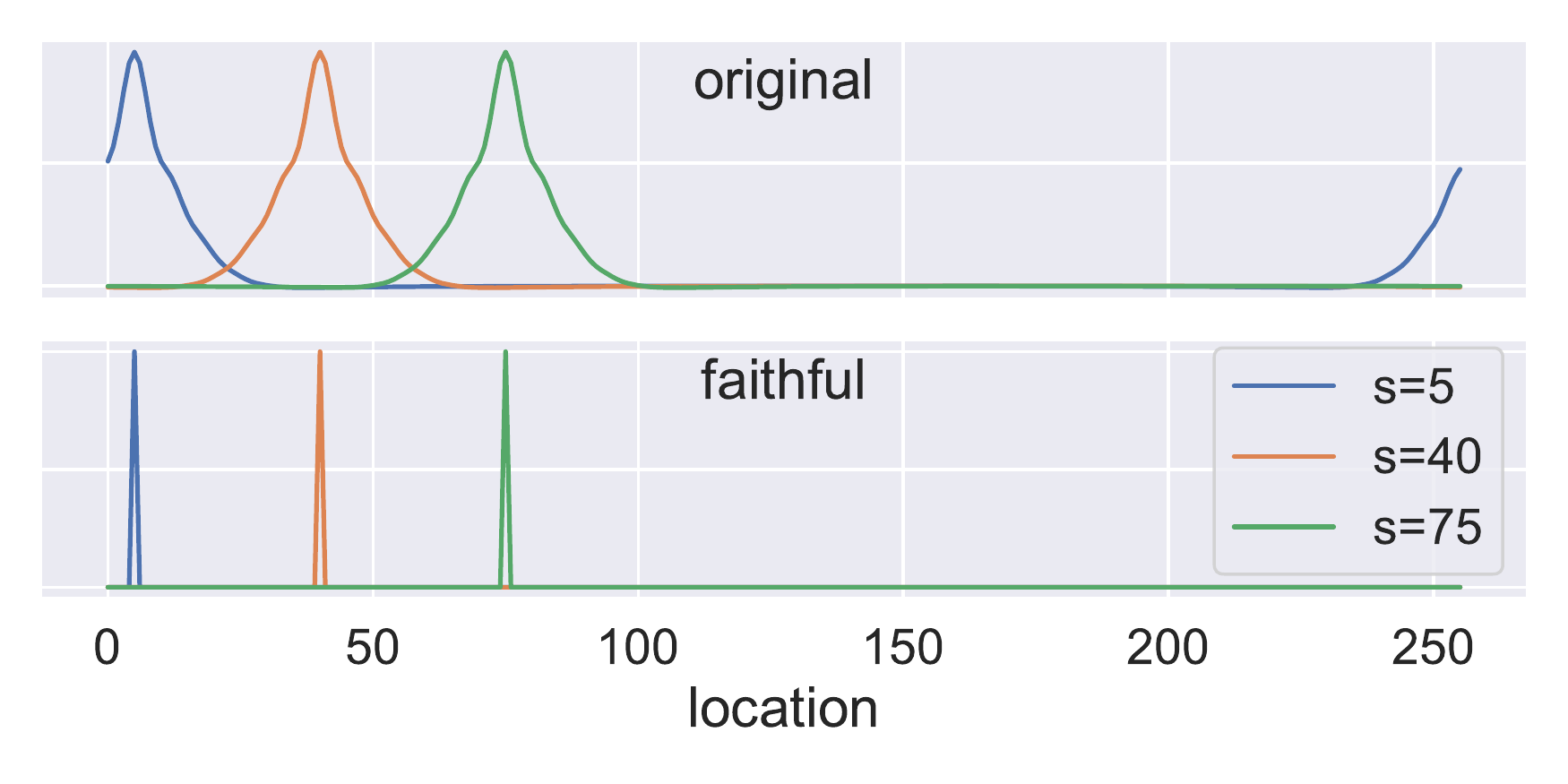}
    \caption{Reconstruction by Algorithm~\ref{algo:reference_reconstruction} for the location function $f(t)=\delta_{t,5},\ \delta_{t,40}, \ \delta_{t,75}$. Perfect reconstruction corresponds to single-peaked spikes at $t=5, 40, 75$, respectively. The broad distributions in the top panel demonstrate a significant loss of information in the original PE. See Section~\ref{sec:DFT_encoding} for the faithful-Encoding (bottom). }
    \label{fig:reconstruction_comparison}
    \vspace{-0.0cm}
\vspace{-0.0cm}
\end{figure}

Figure~\ref{fig:reconstruction_comparison} shows the result of reconstruction. We normalized the modified DFT coefficients so the original $\ell_2$ norm is kept unchanged. All the parameters used are the same as those in Fig.~\ref{fig:frequencyRatio}, i.e.,~$d=256,N_w=80,h=10000, \sigma=4\times \frac{2\pi}{d}$. As expected from the low-pass property, the reconstruction by the original PE failed to reproduce the delta functions. The broad distributions imply that the original PE is not sensitive to the difference in the location up to about 30. As the total sequence length is $N_w=80$, we conclude that the original PE tends to put an extremely strong emphasis on global long-range dependencies within the sequence. 

If the PE is supposed to get a representation of the position function, PE should faithfully reproduce the original position function. Here, we formally define the notion of the faithfulness of PE:

\begin{definition}[Faithfulness of PE]
A positional encoding is said faithful if it is injective (i.e.,~one-to-one) to the position function. 
\end{definition}

For PE, the requirement of faithfulness seems natural. Very interestingly, as long as the sinusoidal bases are assumed, this requirement almost automatically leads to a specific PE algorithm that we call the faithful-Encoding, which is the topic of the next subsection.

\subsection{DFT-based derivation of positional encoding}\label{sec:DFT_encoding}

For the requirement of faithfulness, one straightforward approach is to leverage DFT for positional encoding. The idea is to use the DFT representation as a smooth surrogate for the one-hot function $f_s(t) \triangleq \delta_{s,t}$, where $\delta_{s,t}$ is Kronecker's delta giving 1 only if $s=t$ and 0 otherwise. Let $a_0^{(s)},a_1^{(s)},\ldots, b_1^{(s)}, \ldots, b_K^{(s)}, b_0^{(s)}$ be the DFT coefficients of $f_s(t)$. It is straightforward to compute the coefficients against the real Fourier bases~$\{\varphi_k(t)\}$:
\begin{align}\label{eq:DFT_def_a0}
    a_0^{(s)} &= \sum_{t=0}^{d-1} \delta_{s,t}\varphi_0(t)=\sum_{t=0}^{d-1} \delta_{s,t}\frac{1}{\sqrt{d}}
    =\frac{1}{\sqrt{d}},
    \\ \label{eq:DFT_def_a1}
    a_1^{(s)} &= \sum_{t=0}^{d-1} \delta_{s,t}\varphi_1(t)=\sum_{t=0}^{d-1} \delta_{s,t}\sqrt{\frac{2}{d}}\cos(\omega_1 t) =\sqrt{\frac{2}{d}}\cos(\omega_1 s),
    \\ \nonumber
     &\vdots
     \\ \label{eq:DFT_def_bK}
    b_K^{(s)}&= \sum_{t=0}^{d-1} \delta_{s,t}\varphi_K(t)=\sum_{t=0}^{d-1} \delta_{s,t}\sqrt{\frac{2}{d}}\sin(\omega_K t) =\sqrt{\frac{2}{d}}\sin(\omega_K s),
    \\ \label{eq:DFT_def_b0}
    b_0^{(s)}&=  \sum_{t=0}^{d-1} \delta_{s,t}\varphi_{d-1}(t)=\sum_{t=0}^{d-1} \delta_{s,t}\sqrt{\frac{1}{d}}\cos(\pi t) =\sqrt{\frac{1}{d}}\cos(\pi s),
\end{align}
where $\varphi_k(t)$s are defined by the second equality. Simply using the DFT coefficients, we define the vector representation of the one-hot function as
\begin{align}\label{eq:DFT_def_summary}
\bme^{(s)} \triangleq (a_0^{(s)},a_1^{(s)},\ldots, b_1^{(s)}, \ldots, b_K^{(s)}, b_0^{(s)})^\top,
\end{align}
which is exactly the same as the faithful-Encoding~\eqref{eq:positional_encoding_DFT}.

Because of the general properties of DFT, the following claim is almost evident: 
\begin{theorem}
The DFT-based encoding~\eqref{eq:DFT_def_summary} is faithful. 
\end{theorem}
\begin{proof}
This follows from the existence of inverse transformation in DFT. We can also prove it directly. With Eq.~\eqref{eq:DFT_def_summary}, the r.h.s.~of the DFT expansion
\begin{align}\label{eq:realDFT_general}
    f_s(t) = \frac{1}{\sqrt{d}}[a_0 + b_0 \cos (\pi t)]+  \sqrt{\frac{2}{d}}\sum_{k=1}^K\left( a_k\cos (\omega_k t)
    +b_k \sin (\omega_k t) \right),
\end{align}
is given by
\begin{align*}
    \mbox{r.h.s.}
    &=  \frac{1}{d}[1 + (-1)^{s-t}]
    + \frac{2}{d}\sum_{k=1}^K \cos(\omega_k (s-t))
\end{align*}
It is obvious that the r.h.s.~is 1 if $s=t$. Now, assume $s\neq t$. Using $\cos x =\frac{1}{2}(\rme^{\rmi x} + \rme^{-\rmi x})$, where $\rmi$ is the imaginary unit, and the sum rule of geometric series, we have
\begin{align*}
     \mbox{r.h.s.}&= 
     \frac{1}{d}[1 + (-1)^{s-t}] + \frac{1}{d}
     \frac{c_{s,t} - c_{s,t}^{K+1}  -1 + c_{s,t}^{-K}}{1-c_{s,t}},
\end{align*}
where we have defined $c_{s,t} \triangleq \exp\left(\rmi \frac{2\pi(s-t)}{d}\right)$. By noting $c_{s,t}^{-2K-1} = c_{s,t}$ and $c_{s,t}^{K+1} = (-1)^{s-t}$, it is straightforward to show the r.h.s.~is $0$. Putting all together, the inverse DFT of the faithful-Encoding gives $\delta_{s,t}$, which is the location function.
\end{proof}

In Fig.~\ref{fig:frequencyRatio}, we have shown the distribution of the faithful-Encoding in the Fourier domain. From Eqs.~\eqref{eq:DFT_def_a0}-\eqref{eq:DFT_def_b0}, we see the distribution is given by
\begin{align}
    g_k^{\mathrm{DFT}} = \frac{1}{d}(\delta_{k,0} + \delta_{k,d-1})
    +\frac{2}{d}\sum_{l=1}^K\delta_{k,l},
\end{align}
where $\delta_{k,0}$ etc.~are Kronecker's delta function. This distribution is flat except for the terminal points at $k=0,d-1$. Hence, unlike the original PE, the proposed faithful PE does not have any bias on the choice of the Fourier components. 

With this flat distribution, we also did the reconstruction experiment. The result is shown in the bottom panel in Fig.~\ref{fig:reconstruction_comparison}. We see that the location functions are perfectly reconstructed with the faithful PE.

\section{Experiments}
In this section, we want to demonstrate the ability of DFStrans to detect and diagnose anomalies. To do so, we compare the method with four other baselines algorithms on four datasets. 
\subsection{Datasets and baselines} \label{lab:cs_data_res}



    
 

\paragraph{\textbf{Industrial Case Study}} The industrial dataset used in this study was obtained from a physical model that simulates the behavior of an elevator during up and down journeys. The physical model was designed by a domain expert from a collaborating company that specializes in the manufacturing of elevators. This model was designed to produce fault conditions during these journeys, such as reduced lubrication, misalignment, or localized peaks in the guidance system, or demagnetization or loss of inductance in the electric machine. In this study, three effects related to misalignment, reduced lubrication, and localized peaks in the guidance system were introduced into the model. The first two effects increase the friction between the cabin or counterweight and the rails, and are referred to as \textit{friction journeys (FJs)}. Localized bumps cause local impacts on the cabin, resulting in spike-like deformations, which are referred to as \textit{point anomalous journeys (PAJs)}. Table \ref{tab:elevator} summarizes the sensors installed in the elevator.

\begin{table}[!ht]
\caption{Summary of the sensors installed in the elevator.} 
\label{tab:elevator}
\centering
    \resizebox{0.5\textwidth}{!}{%
    \begin{tabular}{ll}
    \toprule[1.5pt]
    \textbf{Sensor name}                 & \textbf{Description}  \\ \midrule
    \textit{Alpha}                       & Angular acceleration of the pulley                    \\ 
    \textit{Ax, Ay, Az}         & Lateral acceleration of cabin on X, Y, and Z axis     \\ 
    \textit{Fc, Fcw}            & Tension on the cabin's and counterweight cable        \\ 
    \begin{tabular}[c]{@{}c@{}}\textit{FrictionCabin,}\\\textit{FrictionCw}\end{tabular}  & Cabin and counterweight friction \\
    \textit{Fsupport}           & Force on the support of the machine-pulley            \\ 
    \textit{Id, Iq}             & Direct and quadrature power                           \\ 
    \textit{Omega}                       & Angular speed of the pulley                           \\ 
    \textit{Phi}                         & Angular position of the pulley                        \\ 
    \textit{PulleyAz}           & Vertical acceleration of the pulley                   \\ 
    \textit{Vc, Vcw}                     & Cabin and Counterweight speed                         \\ 
    \textit{Vd, Vq}                      & Direct and quadrature voltage                         \\ 
    \textit{Zc, Zcw}                     & Cabin and counterweight position                      \\ \bottomrule[1.5pt]
\end{tabular}%
}
\end{table}

\paragraph{\textbf{Baseline algorithms}} The goal of this work is mainly based on proposing a solution for anomaly diagnosis, but without losing classification performance. Therefore, we compare with some of these state-of-the-art time series classification models. However, these algorithms are not meant for anomaly diagnosis. We compare our classification performance against: (1) \textbf{MH1DCNN-LSTM} \cite{Canizo2019}, (2) \textbf{TapNet} \cite{zhang2020tapnet}, (3) \textbf{InceptionTime} \cite{ismail2020inceptiontime} and (4) \textbf{MLSTM-FCN} \cite{karim2019multivariate}. See Appendix for details on model's hyperparameters.

\paragraph{\textbf{Benchmark datasets}} We evaluated the methods in three other datasets apart from the elevator use-case: both (1) \textbf{Soil Moisture Active Passive satellite (SMAP)} \cite{hundman2018detecting} and (2) \textbf{Mars Science Laborator rover (MSL)} \cite{hundman2018detecting} are public datasets provided by NASA that consist of telemetry data from spacecraft monitoring systems, and (3) \textbf{Server Machine Dataset (SMD)} \cite{su2019robust} is a dataset collected from a large Internet company. These three datasets are unsupervised, so the labeled testing data have been used to validate the models.

\paragraph{\textbf{Data preprocessing and evaluation}} 
Regarding data preprocessing, all data have been scaled (independently for each sensor) between 0 and 1, using MinMaxScaler module of Sklearn \cite{scikit-learn}. For benchmark datasets, we have divided the data into sub-series $\sfU_i$ of $T=500$ points  and labeled them as 1 if an anomaly has occurred during that period and 0 otherwise. For algorithms using multi-head CNNs, these time-series have been divided into non-overlapping windows of length $w_l$, i.e. $\sfU_i = \{\sfU_i^{(1)},...,\sfU_i^{(N_w)}\}$. We have used a 5-fold cross-validation strategy for training and evaluation, using $70\%$ for training, $15\%$ for validation, and $15\%$ for testing. 

\subsection{Anomaly detection} 

\begin{table}[tb]
\caption{Results obtained with the models for the different datasets in terms of Precision (\textbf{P}), Recall (\textbf{R}) and F1-Score (\textbf{F1}). Best scores are highlighted in \textbf{bold}.} 
\label{tab:qresults}
\centering
\resizebox{0.98\textwidth}{!}{%
\begin{tabular}{c|ccccccccccccccc}
\toprule
\multicolumn{1}{l|}{\multirow{3}{*}{\textbf{Datasets}}} &
  \multicolumn{15}{c}{\textbf{Methods}} \\ \cline{2-16} 
\multicolumn{1}{l|}{} &
  \multicolumn{3}{c|}{\textbf{MH1DCNN-LSTM}} &
  \multicolumn{3}{c|}{\textbf{InceptionTime}} &
  \multicolumn{3}{c|}{\textbf{TapNet}} &
  \multicolumn{3}{c|}{\textbf{MLSTM-FCN}} &
  \multicolumn{3}{c}{\textbf{DFStrans }} \\ \cline{2-16} 
\multicolumn{1}{c|}{} &
  P &
  R &
  \multicolumn{1}{c|}{F1} &
  P &
  R &
  \multicolumn{1}{c|}{F1} &
  P &
  R &
  \multicolumn{1}{c|}{F1} &
  P &
  R &
  \multicolumn{1}{c|}{F1} &
  P &
  R &
  F1 \\ \hline
\textbf{Elevator} &
  0.993 &
  0.914 &
  \multicolumn{1}{c|}{0.951} &
   \textbf{1} & 
   0.892 &
  \multicolumn{1}{c|}{0.944} &
   0.398&
   0.711&
  \multicolumn{1}{c|}{0.510} &
  0.913 &
   0.724&
  \multicolumn{1}{c|}{0.808} &
   0.989 &
  \textbf{0.917} &
  \textbf{0.952} \\ 
\textbf{SMD} &
  0.913 &
  0.744 &
  \multicolumn{1}{c|}{0.807} &
   0.856&
   0.925&
  \multicolumn{1}{c|}{0.882} &
   0.842&
   0.606&
  \multicolumn{1}{c|}{0.706} &
   0.951&
   0.853&
  \multicolumn{1}{c|}{0.895} &
  \textbf{0.973} &
  \textbf{0.946} &
 \textbf{ 0.959} \\ 
\textbf{MSL} &
   0.821& 
   0.821&
  \multicolumn{1}{c|}{0.821} &
  \textbf{0.965} &
  \textbf{0.948} &
  \multicolumn{1}{c|}{\textbf{0.956}} &
  0.875 &
  0.867 &
  \multicolumn{1}{c|}{0.857} &
 0.952 &
  0.925 &
  \multicolumn{1}{c|}{0.938} &
   0.867 &
   0.882 &
   0.874 \\ 
\textbf{SMAP} &
   0.923&
   0.805&
  \multicolumn{1}{c|}{0.859} &
  0.853 &
  0.687 &
  \multicolumn{1}{c|}{0.761} &
  0.66 &
  \textbf{0.908} &
  \multicolumn{1}{c|}{0.765} &
  \textbf{0.963} &
  0.833 &
  \multicolumn{1}{c|}{\textbf{0.894}} &
   0.883&
  0.826 & 0.853
   \\ \midrule
   \textbf{mean} &
   0.912&
   0.821&
  \multicolumn{1}{c|}{0.859} &
  0.918 &
  0.863 &
  \multicolumn{1}{c|}{0.888} &
  0.694 &
  0.773 &
  \multicolumn{1}{c|}{0.709} &
  \textbf{0.945} &
  0.834 &
  \multicolumn{1}{c|}{0.884} & 0.931 &
   \textbf{0.893} & \textbf{0.911}
   \\ \bottomrule
\end{tabular}%
}
\end{table}
Our model was evaluated on four datasets using four baseline algorithms. As shown in Table \ref{tab:qresults}, for the elevator use case, DFStrans was the best performer in terms of Recall and F1. Additionally, DFStrans was the top performer on the SMD dataset in all terms. On the MSL dataset, InceptionTime achieved the best results, outperforming the other models. Finally, on the SMAP dataset, MLSTM-FCN performed the best in terms of F1, followed by DFSTrans and MH1DCNN-LSTM. Overall, our algorithm achieved the best average results in terms of Recall and F1, indicating that DFStrans is a competitive algorithm for anomaly detection.  It is worth noting that our algorithm also focuses on diagnosing anomalies rather than providing competitive results for anomaly detection.

\paragraph{\textbf{Ablation on positional encoding}} Here we compare the results obtained using Vaswani's positional encoder and using the faithful-Encoding. As shown in Table \ref{tab:qresults_pe}, the faithful-Encoding achieves better results than Vaswani's positional encoding in three of the four datasets, demonstrating the importance of faithful-Encoding. 

\begin{table}[!ht]
\caption{Results obtained with the models for the different datasets in terms of Precision (\textbf{P}), Recall (\textbf{R}) and F1-Score (\textbf{F1}) under different encoding strategies. Best scores are highlighted in \textbf{bold}.} 
\label{tab:qresults_pe}
\centering
\resizebox{0.6\textwidth}{!}{%
\begin{tabular}{c|cccccc}
\toprule
\multicolumn{1}{l|}{\multirow{3}{*}{\textbf{Datasets}}} &
  \multicolumn{6}{c}{\textbf{Encoding strategies}} \\ \cline{2-7} 
\multicolumn{1}{l|}{} &
  \multicolumn{3}{c|}{\textbf{w Vaswani's encoding }} &
  \multicolumn{3}{c}{\textbf{w faithful-Encoding }} \\ \cline{2-7} 
\multicolumn{1}{c|}{} &
  P &
  R &
  \multicolumn{1}{c|}{F1} &
  P &
  R &
  F1 \\ \hline
\textbf{Elevator} &
  0.952 &
  0.914 &
  \multicolumn{1}{c|}{0.931} &
  \textbf{0.989} &
  \textbf{0.917} &
  \textbf{0.952} \\ 
\textbf{SMD} &
  \textbf{0.986} &
  0.942 &
  \multicolumn{1}{c|}{0.963} &
  0.985 &
  \textbf{0.948} &
  \textbf{0.965} \\ 
\textbf{MSL} &
   0.862&
   0.838&
  \multicolumn{1}{c|}{0.849} &
   \textbf{0.867}&
   \textbf{0.882}&
   \textbf{0.874}\\ 
\textbf{SMAP} &
   \textbf{0.887}&
   0.823&
  \multicolumn{1}{c|}{\textbf{0.854}} &
   0.883&
  \textbf{ 0.826} & 0.853
   \\ \midrule
   \textbf{mean} &
   0.922&
  0.879&
  \multicolumn{1}{c|}{0.899} &
   \textbf{0.931}&
   \textbf{0.893} & \textbf{0.911}
   \\ \bottomrule
\end{tabular}%
}
\end{table}

\subsection{Anomaly diagnosis} 

Next, we evaluate the capability in anomaly diagnosis using the elevator's use case, in which expert feedback is available. Here, we show a few examples of how the temporal and spatial attention matrices, $\sfA$ and $\sfB$ respectively (defined in Section \ref{sec:a_b_mat}), are used to obtain intrinsic interpretability for diagnosing the detected anomalies. 

\begin{figure*}[!ht]
    \centering
    \begin{subfigure}[b]{0.49\textwidth}
        \centering
        \includegraphics[width=0.8\textwidth]{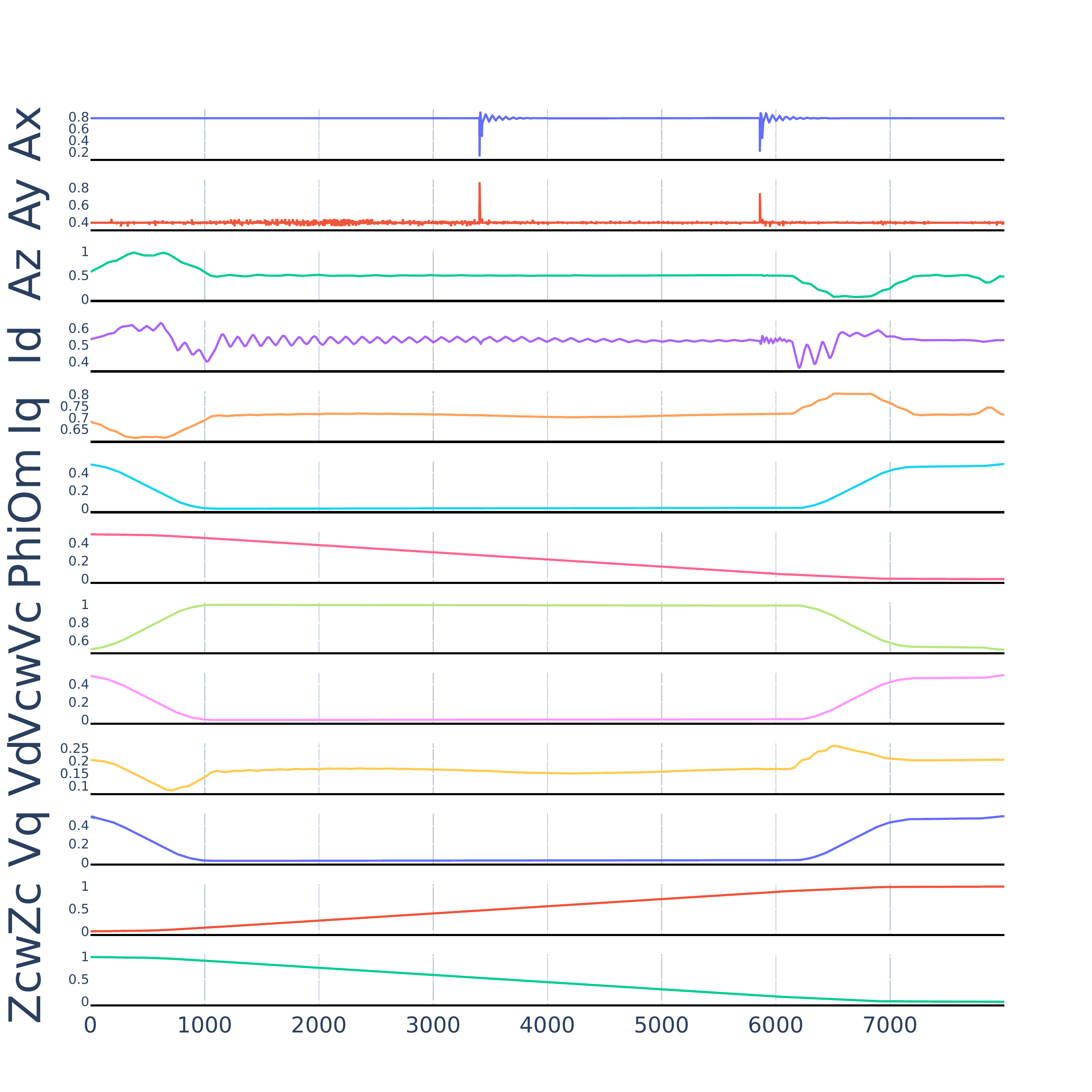}
        \subcaption{}
        \label{fig:PAJ38}
    \end{subfigure}\hfill
    \begin{subfigure}[b]{0.49\textwidth}
        \centering
        \includegraphics[width=0.8\textwidth]{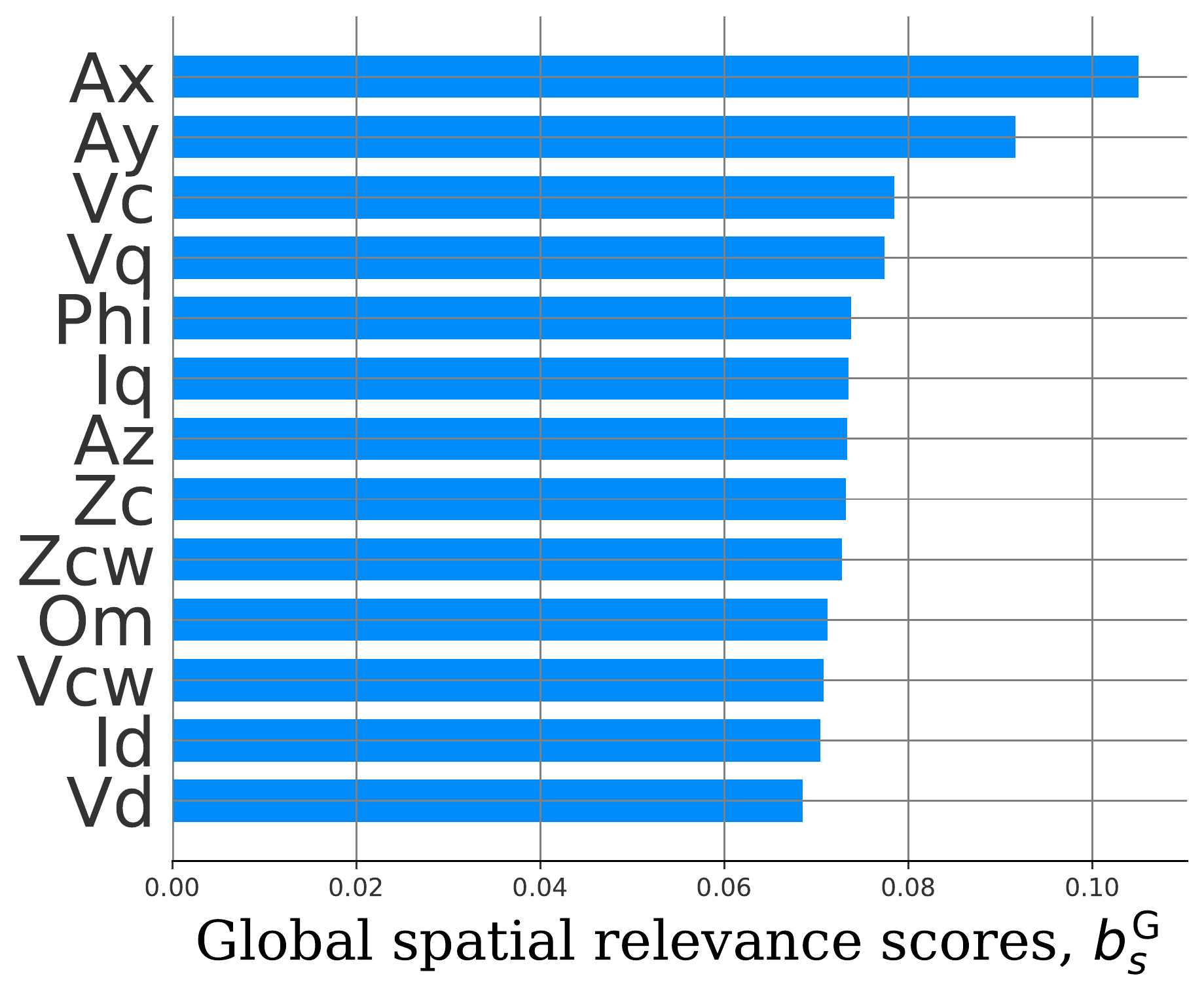}
        \subcaption{}
        \label{fig:GSRPAJ38}
    \end{subfigure}
    \\
    \begin{subfigure}[b]{0.3\textwidth}
        \centering
        \includegraphics[width=\textwidth]{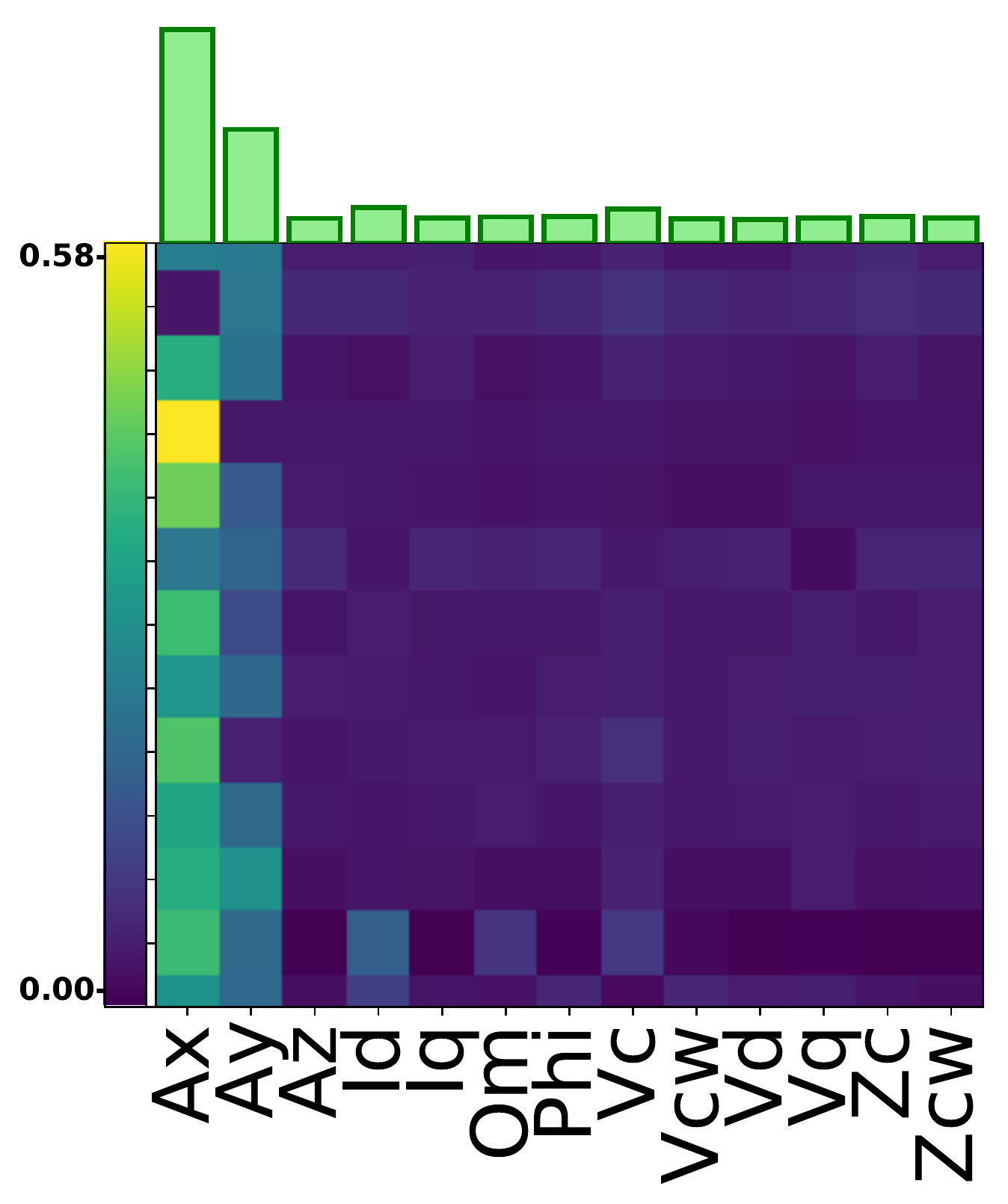}
        \subcaption{}
        \label{fig:lsa_t34}
    \end{subfigure}\hfill
    \begin{subfigure}[b]{0.3\textwidth}
        \centering
        \includegraphics[width=\textwidth]{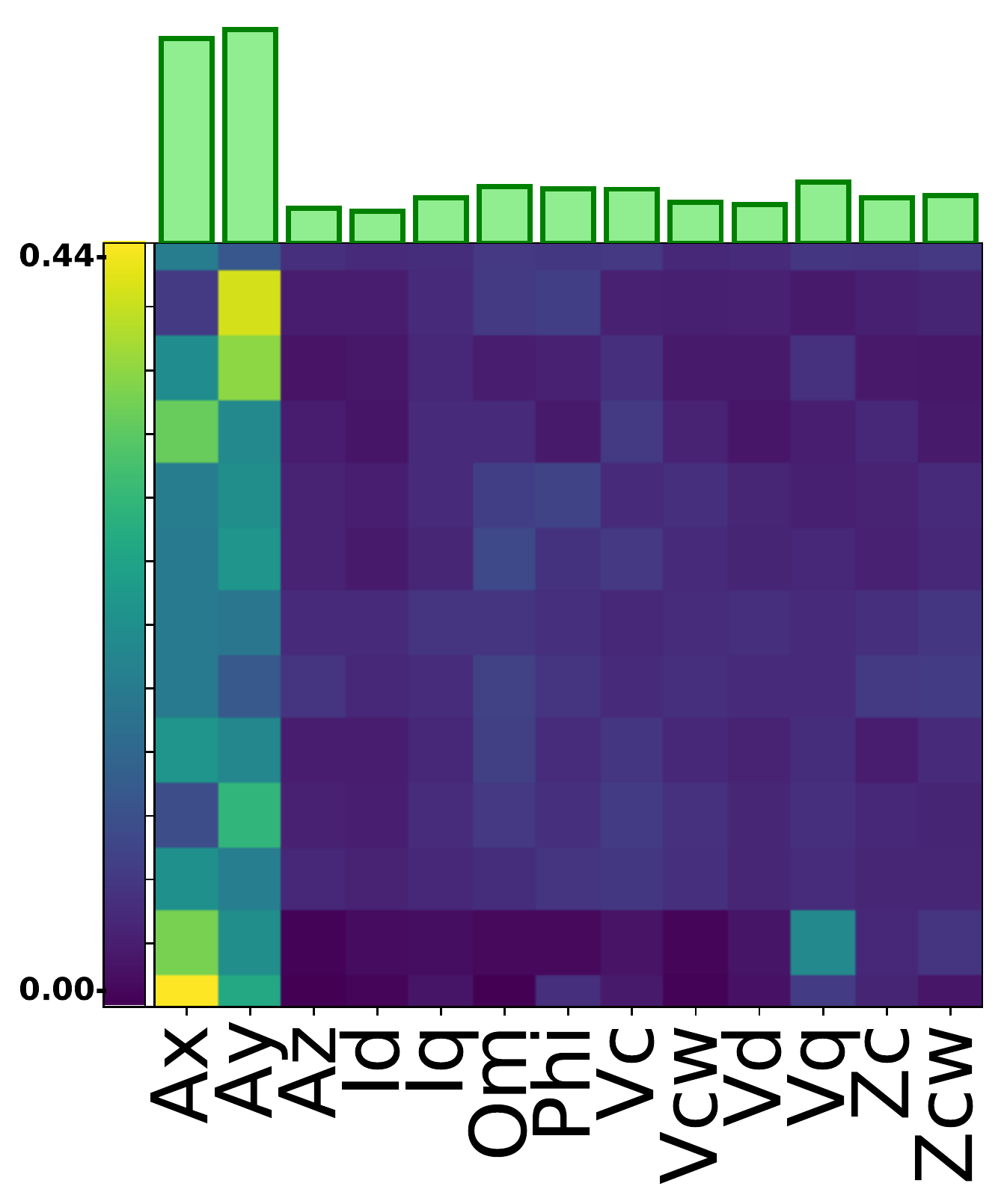}
        \subcaption{}
        \label{fig:lsa_t58}
    \end{subfigure}\hfill
    \begin{subfigure}[b]{0.3\textwidth}
        \centering
        \includegraphics[width=\textwidth]{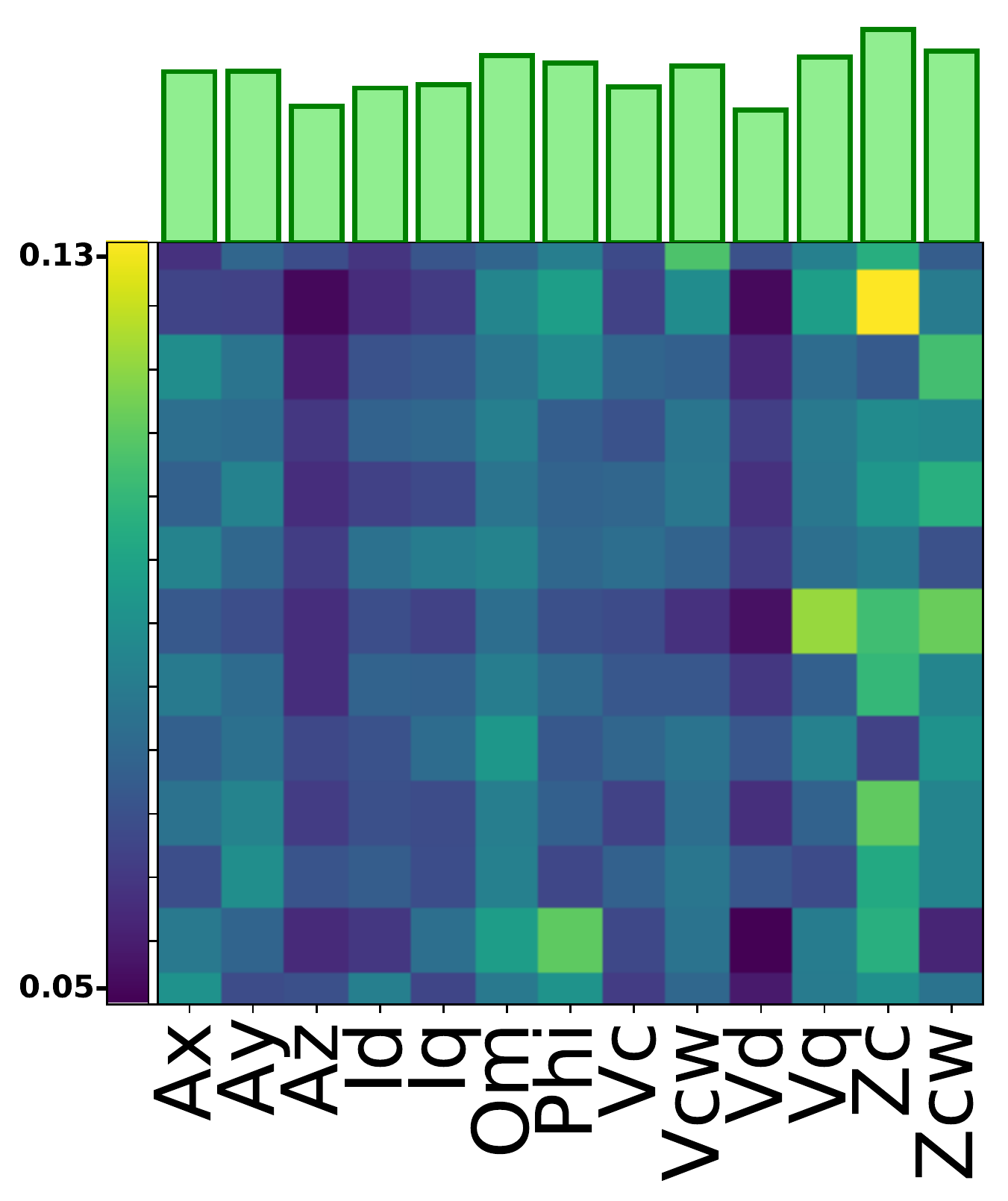}
        \subcaption{}
        \label{fig:lsa_t75}
    \end{subfigure}

    \caption{The first plot \textbf{(a)} shows a PAJ containing two point anomalies, at $\tau=34$ and $\tau=58$, which can be seen as ripples in the accelerations. Then, the plot \textbf{(b)} shows the global spatial relevances. Plots \textbf{(c)}, \textbf{(d)} and \textbf{(e)} show the spatial attention matrix and the spatial relevance vectors for time steps $\tau=34$, $\tau=58$ and $\tau=75$, respectively.}
    \label{fig:lsa}
\end{figure*}

In Figure \ref{fig:lsa}, we present the spatial attention matrix and spatial relevance vector for a PAJ containing two point anomalies, as identified by a domain expert, at time segments $\tau=34$ and $\tau=58$, reflected in the accelerations $Ax$ and $Ay$. Only a subset of sensors is shown for space limitation. The local spatial interpretations allow us to investigate the sensor that the model is attending to at each time segment. We selected the time segments where the anomalies occur, $\tau=34$ and $\tau=58$, and plotted the spatial attention matrix $\sfB^{(\tau)}$ with the relevance scores $b_s^{(\tau)}$ of each sensor $s$ to examine whether the model is focusing on the accelerations. We also randomly selected another time segment, $\tau=75$, where there is no anomaly, to examine the behavior of the spatial attention at that point. In time segments $\tau=34$ and $\tau=58$ (Figures \ref{fig:lsa_t34} and \ref{fig:lsa_t58}), we observe that the relevance of the accelerations $Ax$ and $Ay$ is high while the relevance of the other sensors is low, as expected. In contrast, for time segment $\tau=75$, all sensors have similar relevance, and the model does not focus on any particular sensor. We also observe that the global spatial relevances are high for both accelerations.

\begin{figure*}[!ht]
    \begin{minipage}[b]{0.34\textwidth}
        \centering
        \includegraphics[width=\textwidth,height=8cm]{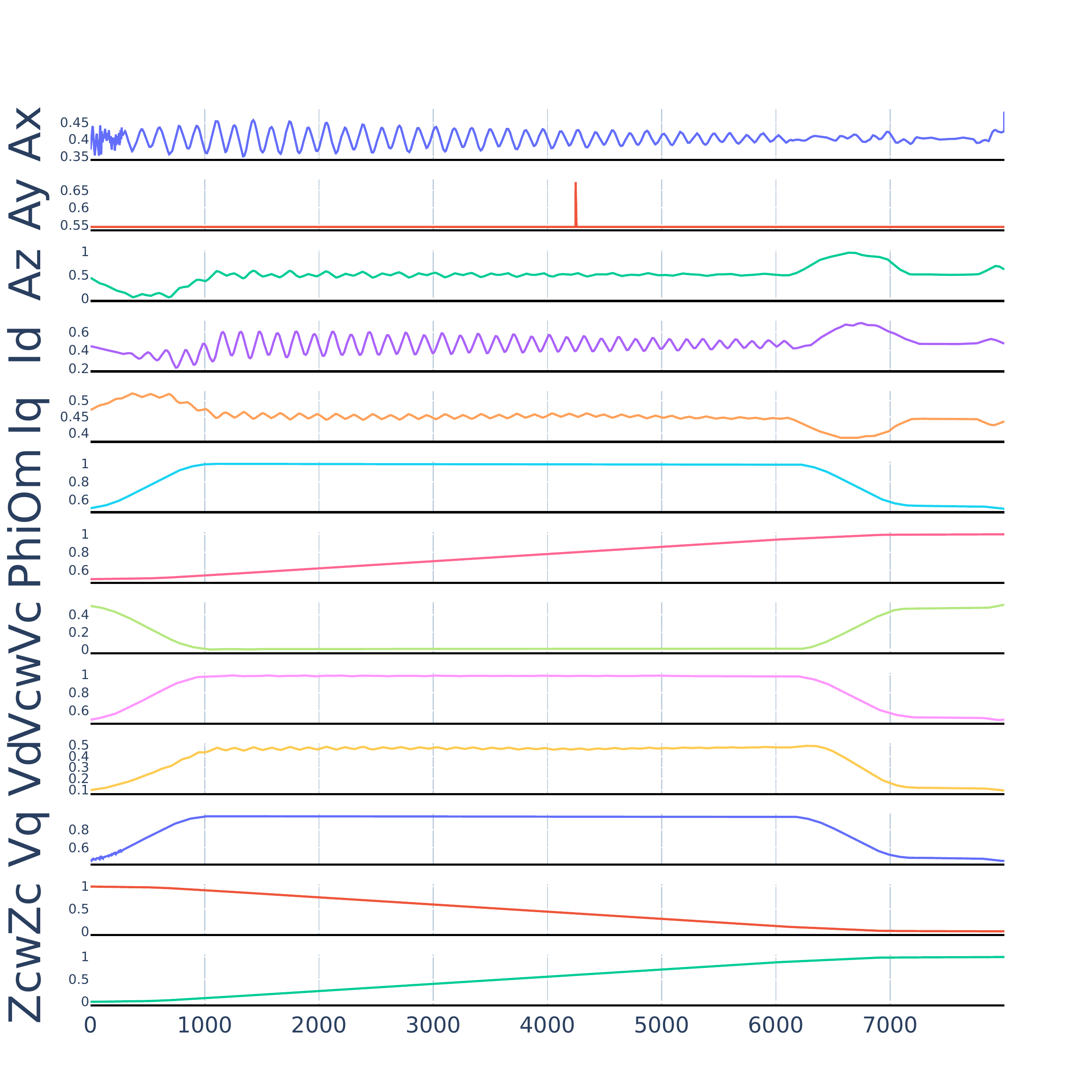}
            \subcaption{}
        \label{fig:mix1}
    \end{minipage}
    \begin{minipage}[b]{0.64\textwidth}
        \centering
        \begin{subfigure}[b]{0.49\textwidth}
            \centering
            \includegraphics[width=3.5cm]{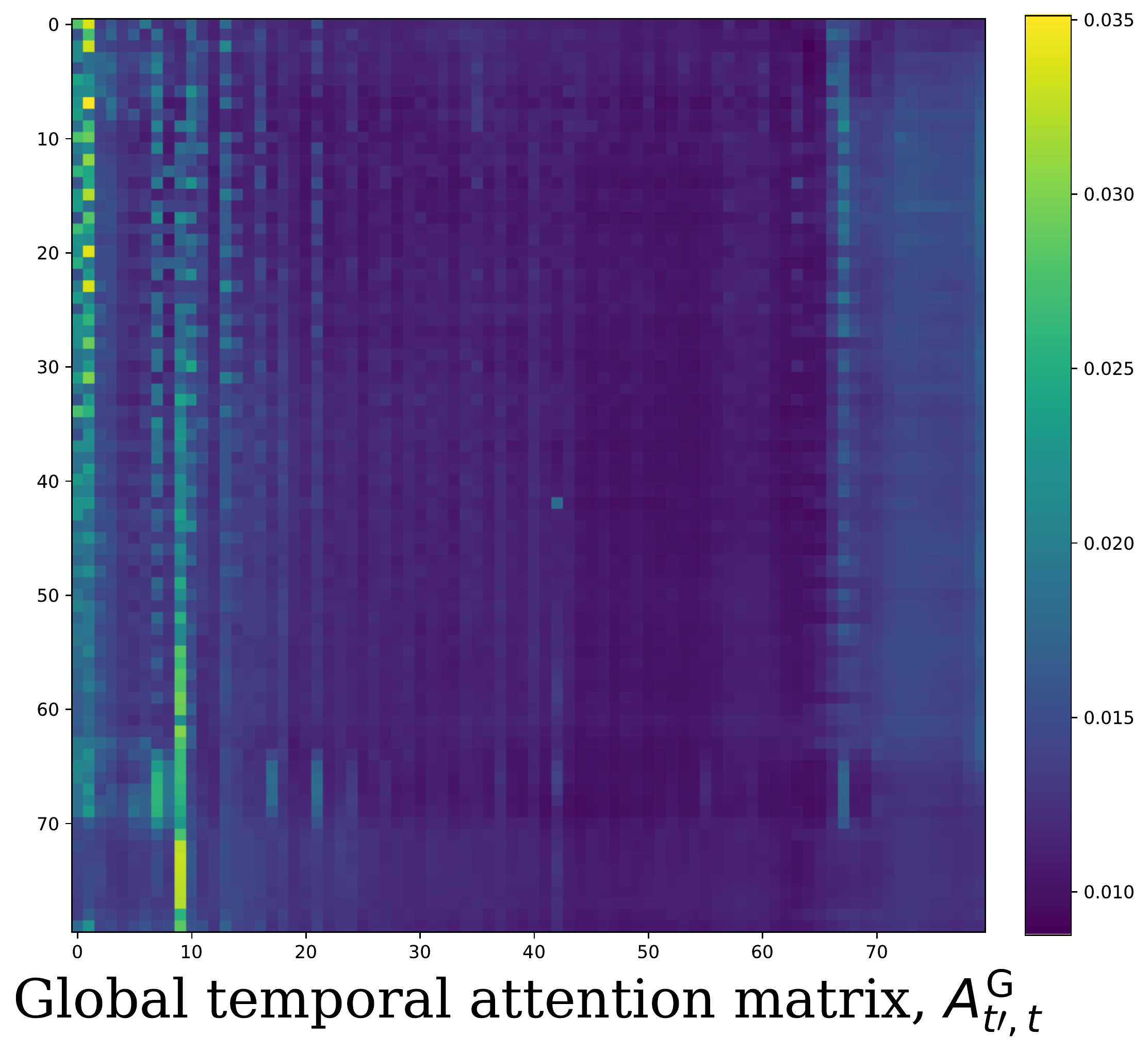}
            \subcaption{}
            \label{fig:mix2}
        \end{subfigure}
        \centering
        \begin{subfigure}[b]{0.49\textwidth}
            \centering
            \includegraphics[width=3.5cm,height=3.2cm]{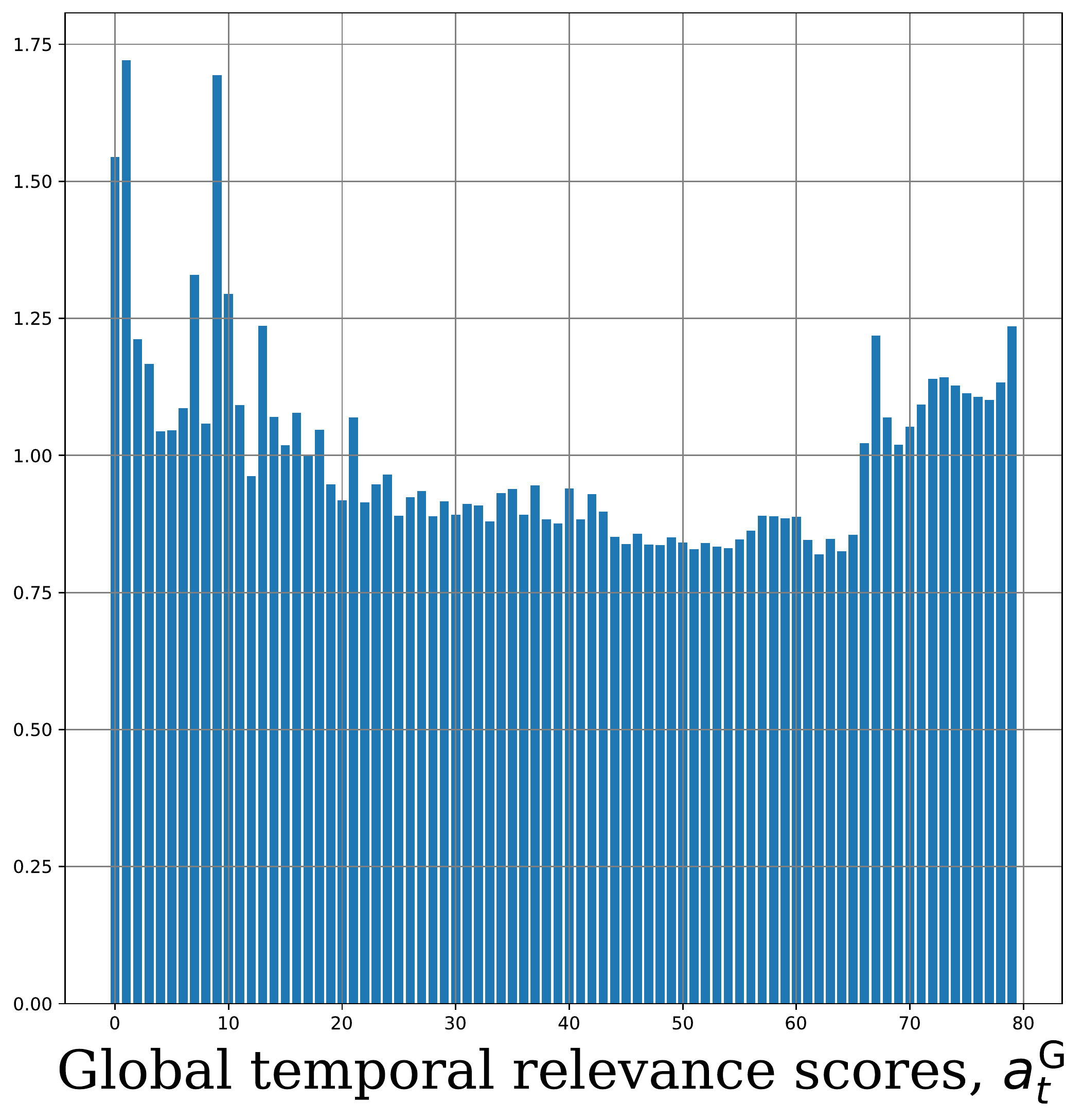}
            \subcaption{}
            \label{fig:mix3}
        \end{subfigure}
        \newline
        \centering
        \begin{subfigure}[b]{0.49\textwidth}
            \centering
            \includegraphics[width=3.5cm]{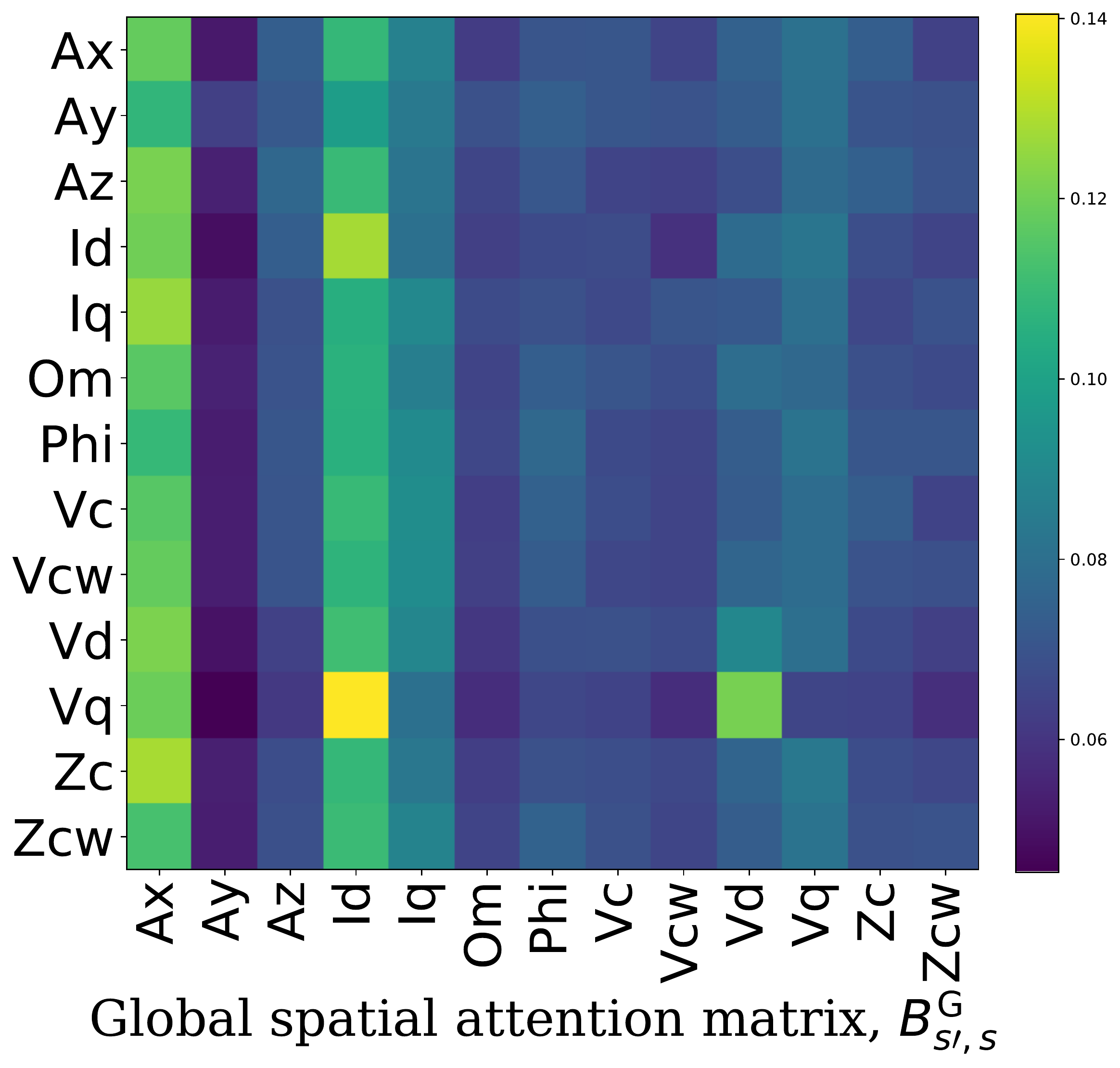}
            \subcaption{}
            \label{fig:mix4}
        \end{subfigure}
        \centering
        \begin{subfigure}[b]{0.49\textwidth}
            \centering
            \includegraphics[width=3.5cm,height=3.4cm]{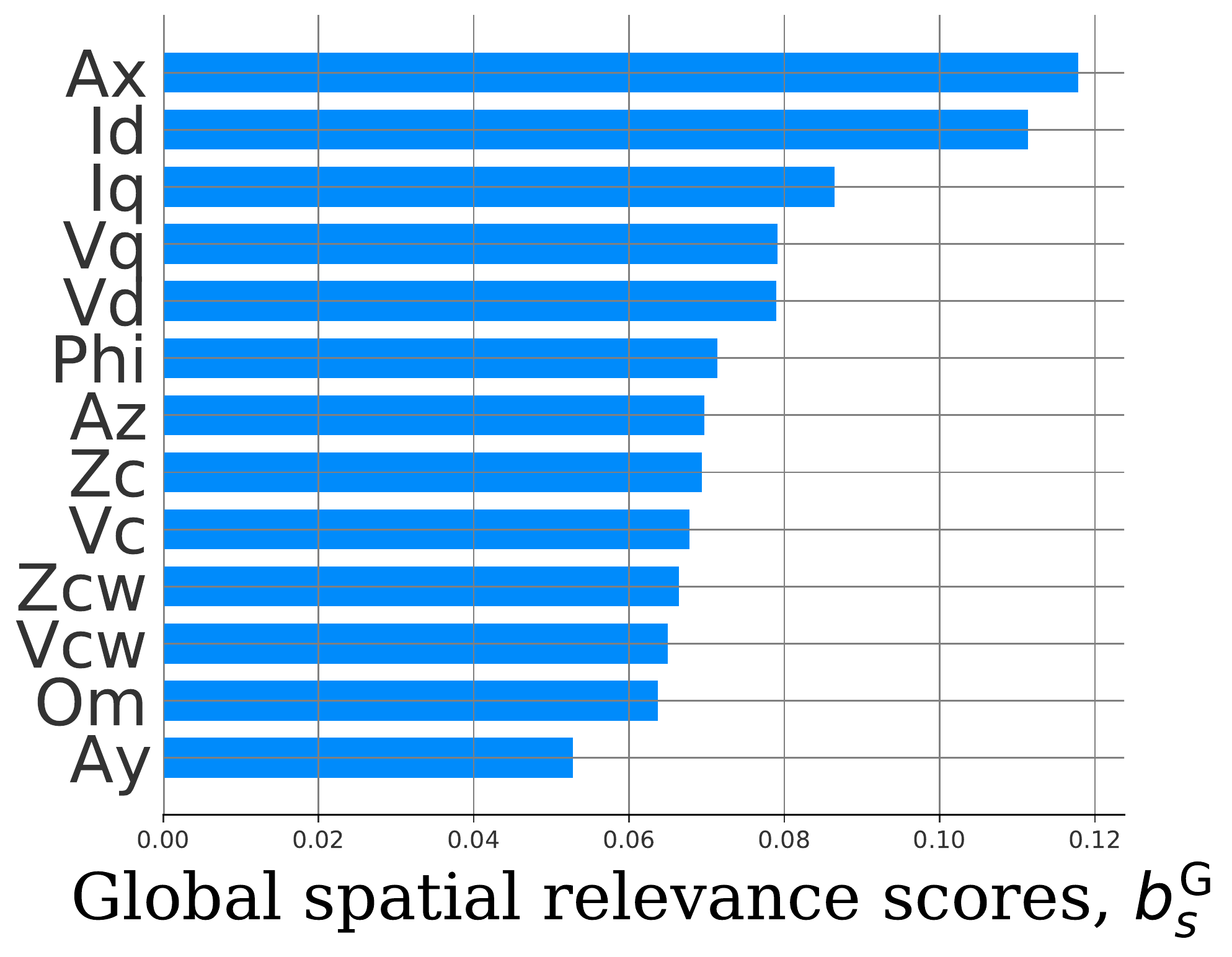}
            \subcaption{}
            \label{fig:mix5}
        \end{subfigure}
    \end{minipage}

    \caption{\textbf{(a)} shows a FJ. Then, \textbf{(b)} shows the global temporal attention matrix, \textbf{(c)} the global temporal attention relevances, \textbf{(d)} the global spatial attention matrix and \textbf{(e)} the global spatial attention relevances.}
    \label{fig:mix}
\end{figure*}
Unlike in PAJs, in FJs, friction is usually present during the whole journey. Therefore, the attention should be somewhat distributed among all time segments, with potentially stronger attention during acceleration and deceleration phases according to the expert's observations. This is demonstrated in Figure \ref{fig:mix}, where the global temporal attention matrix and global temporal attention relevances show that attention is present throughout the entire journey, with slight peaks at the beginning (acceleration) and end (deceleration) of the journey, rather than any individual time segment standing out significantly as in PAJs. Additionally, FJs tend to increase the required motor torque and electrical consumption, as reflected in the sensors measuring electrical current (i.e., $Iq$ and $Id$) and the electric machine voltages (i.e., $Vq$ and $Vd$). The global spatial attention matrix and global spatial relevance scores in Figure \ref{fig:mix} show that these sensors had the greatest influence on the model's classification of the journey as anomalous.

\begin{figure*}[!ht]
     \centering
     \begin{subfigure}[b]{0.45\textwidth}
         \centering
         \includegraphics[width=\textwidth]{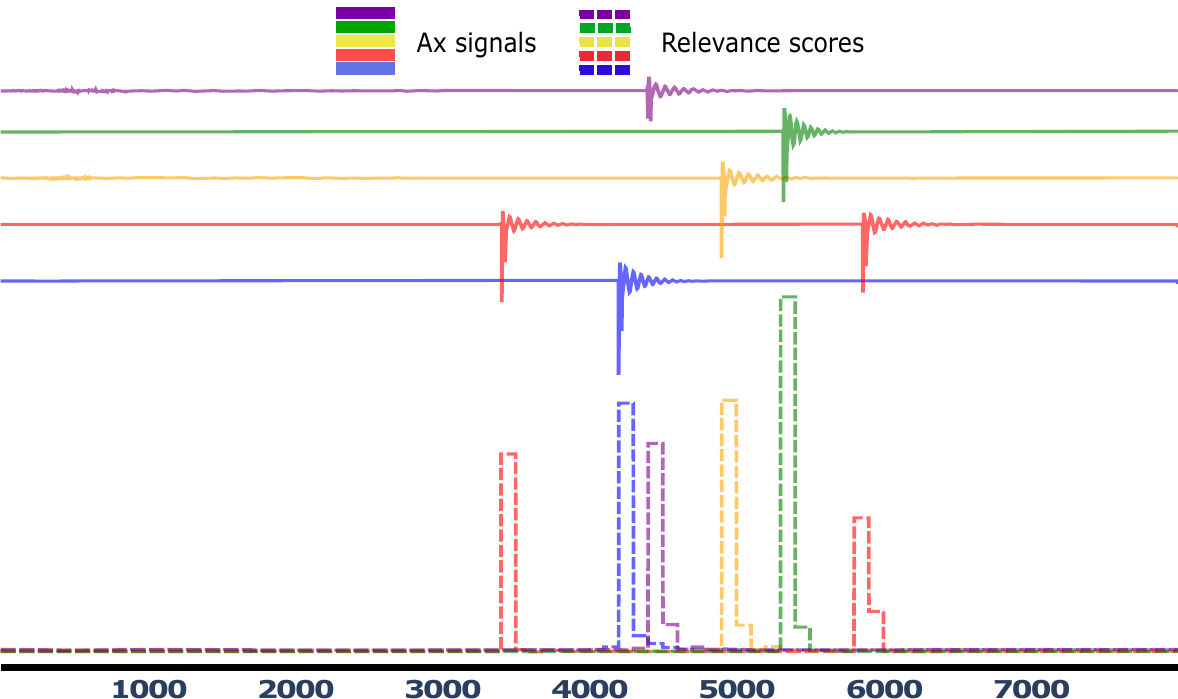}
        \subcaption{}
         \label{fig:paj_ax_local}
     \end{subfigure}
     \hfill
     \begin{subfigure}[b]{0.45\textwidth}
         \centering
         \includegraphics[width=\textwidth]{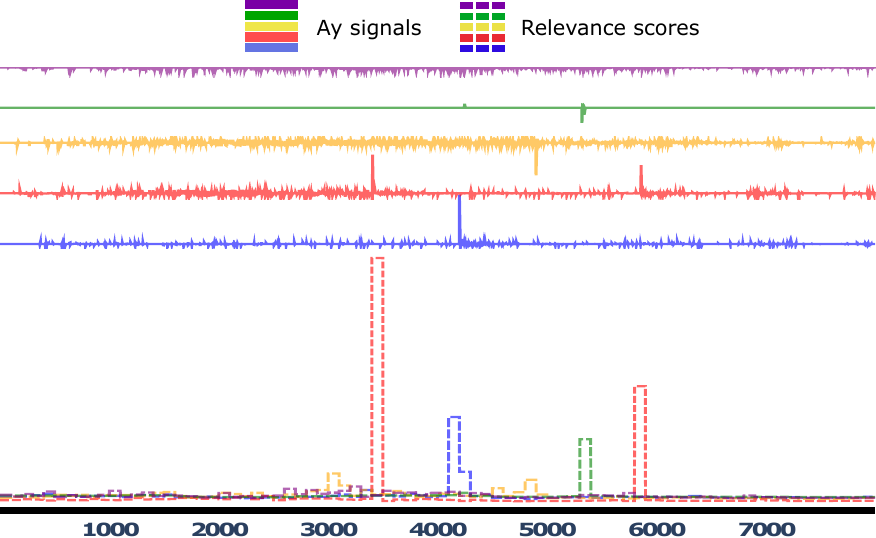}
        \subcaption{}
         \label{fig:paj_ay_local}
     \end{subfigure}
     \hfill
     \begin{subfigure}[b]{0.8\textwidth}
         \centering
         \includegraphics[width=\textwidth]{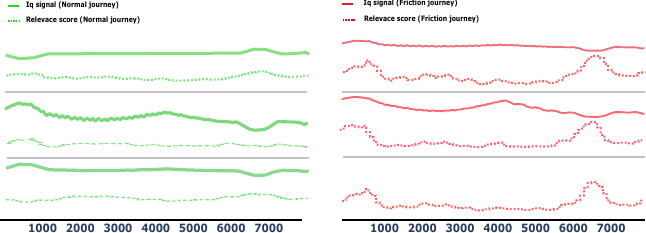}
        \subcaption{}
         \label{fig:paj_id_local}
     \end{subfigure}

      \caption{Local temporal attention plots. Figures \textbf{(a)} and \textbf{(b)} are plots of five PAJs, in which the observations of the $Ax$ and $Ay$ sensors are shown together with their temporal relevance scores $a^{(s)}$. Figure \textbf{(c)} is a comparison of the temporal relevance scores for sensor $Iq$ between three FJs (\textcolor{red}{red} lines) and three normal journeys (\textcolor{green}{green} lines).}
    \label{fig:lta}

\end{figure*}
Figure \ref{fig:lta} presents the temporal attention matrix and spatial relevance vectors. The first two plots depict the local temporal relevances for five different PAJs. Accelerations are plotted because, according to the domain expert, anomalies in PAJs are typically reflected in these sensors. Figure \ref{fig:paj_id_local} shows the temporal attention of the Iq sensor on six different journeys, three of which are normal (shown in green) and three of which are FJs with the \textit{Iq} sensor in the top positions (shown in red). In these cases, the distinction between anomalous and non-anomalous journeys is most apparent during the acceleration and deceleration phases, where the attention is stronger in the FJs. Additionally, the temporal relevance is evenly distributed throughout the entire journey.

\paragraph{\textbf{Evaluating Attention}}

 To quantify the trustworthiness of the attention mechanisms, we have used the \textit{AT-Score} metric under different percentages of top-\textit{k} attention weights set to zero. Since the PAJs exhibit anomalous behavior at specific time steps during the journey and the FJs exhibit anomalous behavior throughout the entire time series, we have distinguished both to conduct this experiment.
\begin{figure*}[!ht]
     \centering
     \begin{subfigure}[b]{0.49\textwidth}
         \centering
         \includegraphics[width=0.8\textwidth]{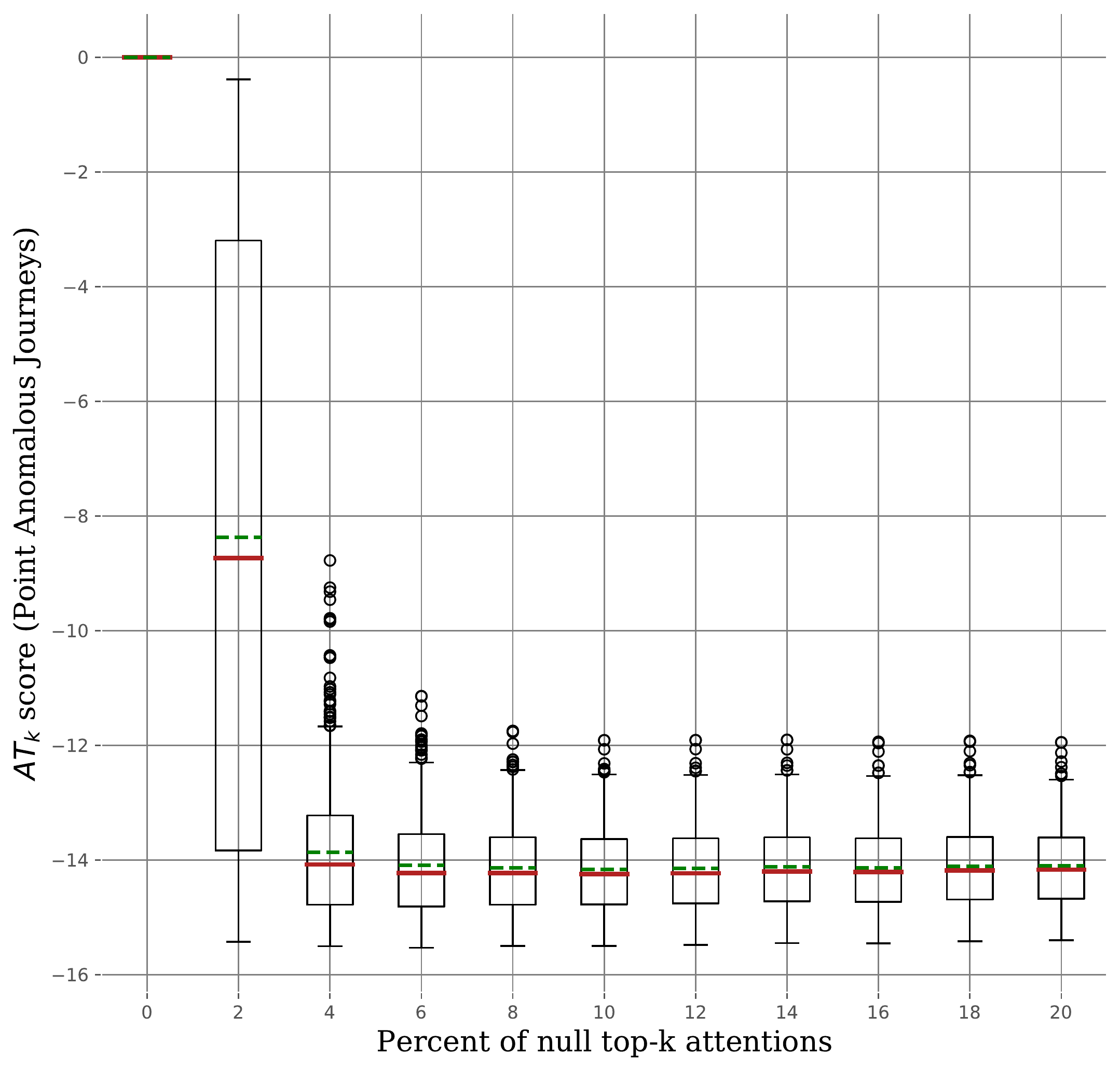}
         \subcaption{}
         \label{fig:PAJ_AT}
     \end{subfigure}\hfill
     \begin{subfigure}[b]{0.49\textwidth}
         \centering
         \includegraphics[width=0.8\textwidth]{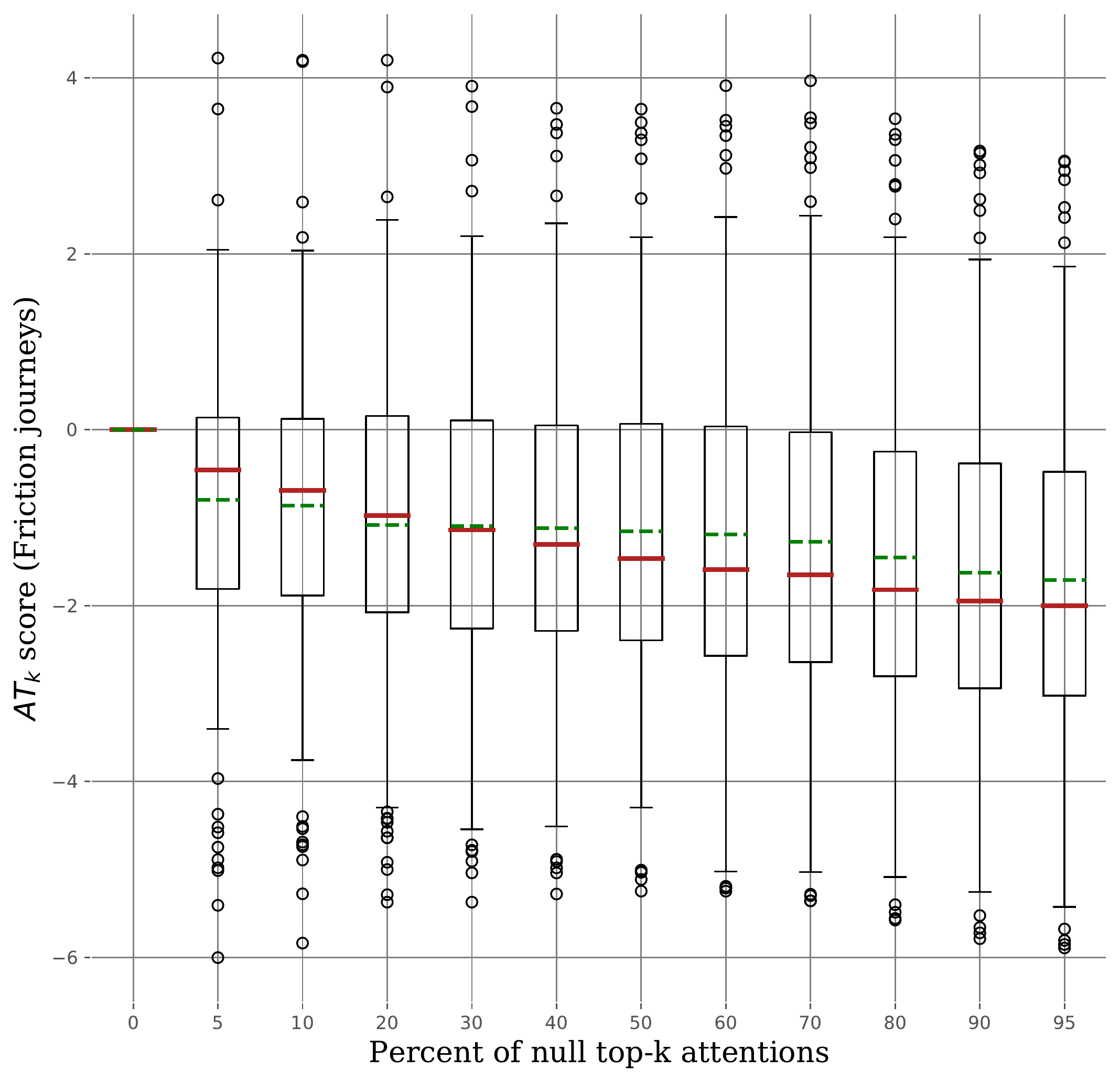}
        \subcaption{}
        \label{fig:FJ_AT}
     \end{subfigure}

    \caption{Boxplots of the AT-Scores obtained under different percentages of top-\textit{k} attention weights set to zero. Plot \textbf{(a)} corresponds to PAJs and plot \textbf{(b)} to corresponds FJs. Median and mean values are represented with \textcolor{red}{red} and \textcolor{green}{green} lines, respectively.}
    \label{fig:at_scores}
    
\end{figure*}

Figure \ref{fig:PAJ_AT} shows that as the weights with the highest attention values in the PAJs are set to zero, the value of the model output decreases. This trend can be seen in the medians and means of these values, represented as red and green lines. This means that the PAJs are classified as normal journeys if the attention mechanism does not focus on the most relevant time-segments, which means that it was previously focusing on where the anomalous behaviour occurred during the journey. However, as shown in Figure \ref{fig:FJ_AT}, this effect is less accentuated and more variable in FJs, because this type of anomalous behavior can be noticed throughout the whole journey.

\section{Conclusions} \label{sec:cncls}
This research presents a new framework that leverages a spatio-temporal dependency model for diagnosing anomalies in noisy multivariate sensor data. The framework combines multi-head 1D CNNs and a Transformer-like spatio-temporal architecture. The multi-head 1D CNNs are used to obtain rich embeddings from raw sensor data while preserving the temporal nature of the data. Then, the Transformer-like architecture allows learning the spatial and temporal dependencies between these embeddings. Additionally, the study mathematically demonstrates the limitations of the positional encoding used in vanilla Transformers, which disproportionately emphasizes long-range correlations and suppresses short- and medium-term location differences, which can be detrimental in anomaly detection scenarios where small changes over time are important. To address this issue, we propose a DFT-based positional encoding, called faithful encoding, that does not introduce any bias in the information about the location of the items and has a theoretical guarantee of faithfulness.

The effectiveness of the proposed framework was demonstrated through evaluations on four different datasets and an ablation study comparing the results using vanilla and faithful positional encodings. The results showed that the faithful encoding improved upon the vanilla encoding in most benchmark datasets, and a sensitivity analysis demonstrated the reliability of attention in diagnosing anomalies. Overall, the proposed DFSTrans framework offers reliable solutions for detecting and diagnosing anomalies in multi-sensor industrial scenarios.

\section*{Acknowledgements}
Jokin Labaien, Xabier De Carlos and Ekhi Zugasti are supported by the DREMIND project of the Basque Government under Grant KK-2022/00049 from the ELKARTEK program. Ekhi Zugasti is part of the Intelligent Systems for Industrial Systems research group of Mondragon Unibertsitatea (IT1676-22), supported by the Department of Education, Universities and Research of the Basque Country. Jokin Labaien, Xabier De Carlos and Ekhi Zugasti are part of the "Intelligent Digital Platforms" IBAT collaboration team between Ikerlan and Mondragon Goi Eskola Politeknikoa. Part of the research was done during Jokin's visit at IBM Research with Pin-Yu Chen and Tsuyoshi Ide.

\bibliographystyle{apalike}
\bibliography{DFStrans}

\begin{thebibliography}{}

\bibitem[Aksan et~al., 2020a]{aksan2020attention}
Aksan, E., Cao, P., Kaufmann, M., and Hilliges, O. (2020a).
\newblock Attention, please: A spatio-temporal transformer for {3D} human
  motion prediction.
\newblock {\em arXiv preprint arXiv:2004.08692}, 2.

\bibitem[Aksan et~al., 2020b]{aksan2020spatio}
Aksan, E., Cao, P., Kaufmann, M., and Hilliges, O. (2020b).
\newblock A spatio-temporal transformer for {3D} human motion prediction.
\newblock {\em arXiv preprint arXiv:2004.08692}.

\bibitem[Aksan et~al., 2021]{aksan2021spatio}
Aksan, E., Kaufmann, M., Cao, P., and Hilliges, O. (2021).
\newblock A spatio-temporal transformer for 3d human motion prediction.
\newblock In {\em 2021 International Conference on 3D Vision (3DV)}, pages
  565--574. IEEE.

\bibitem[An and Cho, 2015]{an2015variational}
An, J. and Cho, S. (2015).
\newblock Variational autoencoder based anomaly detection using reconstruction
  probability.
\newblock {\em Special Lecture on IE}, 2(1):1--18.

\bibitem[Bishop, 2006]{Bishop}
Bishop, C.~M. (2006).
\newblock {\em Pattern Recognition and Machine Learning}.
\newblock Springer-Verlag.

\bibitem[Bl{\'a}zquez-Garc{\'\i}a et~al., 2021]{blazquez2021review}
Bl{\'a}zquez-Garc{\'\i}a, A., Conde, A., Mori, U., and Lozano, J.~A. (2021).
\newblock A review on outlier/anomaly detection in time series data.
\newblock {\em ACM Computing Surveys (CSUR)}, 54(3):1--33.

\bibitem[Buchholz and Jug, 2022]{buchholz2022fourier}
Buchholz, T.-O. and Jug, F. (2022).
\newblock Fourier image transformer.
\newblock In {\em Proceedings of the IEEE/CVF Conference on Computer Vision and
  Pattern Recognition}, pages 1846--1854.

\bibitem[Cai et~al., 2020]{cai2020traffic}
Cai, L., Janowicz, K., Mai, G., Yan, B., and Zhu, R. (2020).
\newblock Traffic transformer: Capturing the continuity and periodicity of time
  series for traffic forecasting.
\newblock {\em Transactions in GIS}, 24(3):736--755.

\bibitem[Canizo et~al., 2019]{Canizo2019}
Canizo, M., Triguero, I., Conde, A., and Onieva, E. (2019).
\newblock {Multi-head CNN–RNN for multi-time series anomaly detection: An
  industrial case study}.
\newblock {\em Neurocomputing}, 363:246--260.

\bibitem[Chen et~al., 2021a]{chen2021nast}
Chen, K., Chen, G., Xu, D., Zhang, L., Huang, Y., and Knoll, A. (2021a).
\newblock Nast: Non-autoregressive spatial-temporal transformer for time series
  forecasting.
\newblock {\em arXiv preprint arXiv:2102.05624}.

\bibitem[Chen et~al., 2021b]{chen2021learning}
Chen, Z., Chen, D., Zhang, X., Yuan, Z., and Cheng, X. (2021b).
\newblock Learning graph structures with transformer for multivariate time
  series anomaly detection in iot.
\newblock {\em IEEE Internet of Things Journal}.

\bibitem[Doshi et~al., 2022]{doshi2022reward}
Doshi, K., Abudalou, S., and Yilmaz, Y. (2022).
\newblock Reward once, penalize once: Rectifying time series anomaly detection.
\newblock In {\em 2022 International Joint Conference on Neural Networks
  (IJCNN)}, pages 1--8. IEEE.

\bibitem[Grigsby et~al., 2021]{grigsby2021long}
Grigsby, J., Wang, Z., and Qi, Y. (2021).
\newblock Long-range transformers for dynamic spatiotemporal forecasting.
\newblock {\em arXiv preprint arXiv:2109.12218}.

\bibitem[Huang et~al., 2022]{huang2022spatial}
Huang, L., Mao, F., Zhang, K., and Li, Z. (2022).
\newblock Spatial-temporal convolutional transformer network for multivariate
  time series forecasting.
\newblock {\em Sensors}, 22(3):841.

\bibitem[Huang et~al., 2020]{huang2020hitanomaly}
Huang, S., Liu, Y., Fung, C., He, R., Zhao, Y., Yang, H., and Luan, Z. (2020).
\newblock Hitanomaly: Hierarchical transformers for anomaly detection in system
  log.
\newblock {\em IEEE Transactions on Network and Service Management},
  17(4):2064--2076.

\bibitem[Hundman et~al., 2018]{hundman2018detecting}
Hundman, K., Constantinou, V., Laporte, C., Colwell, I., and Soderstrom, T.
  (2018).
\newblock Detecting spacecraft anomalies using lstms and nonparametric dynamic
  thresholding.
\newblock In {\em Proceedings of the 24th ACM SIGKDD international conference
  on knowledge discovery \& data mining}, pages 387--395.

\bibitem[Ioffe and Szegedy, 2015]{ioffe2015batch}
Ioffe, S. and Szegedy, C. (2015).
\newblock Batch normalization: Accelerating deep network training by reducing
  internal covariate shift.
\newblock In {\em International conference on machine learning}, pages
  448--456. PMLR.

\bibitem[Ismail~Fawaz et~al., 2020]{ismail2020inceptiontime}
Ismail~Fawaz, H., Lucas, B., Forestier, G., Pelletier, C., Schmidt, D.~F.,
  Weber, J., Webb, G.~I., Idoumghar, L., Muller, P.-A., and Petitjean, F.
  (2020).
\newblock {Inceptiontime}: Finding {AlexNet} for time series classification.
\newblock {\em Data Mining and Knowledge Discovery}, 34(6):1936--1962.

\bibitem[Karim et~al., 2019]{karim2019multivariate}
Karim, F., Majumdar, S., Darabi, H., and Harford, S. (2019).
\newblock Multivariate {LSTM-FCNs} for time series classification.
\newblock {\em Neural Networks}, 116:237--245.

\bibitem[Kazemi et~al., 2019]{kazemi2019time2vec}
Kazemi, S.~M., Goel, R., Eghbali, S., Ramanan, J., Sahota, J., Thakur, S., Wu,
  S., Smyth, C., Poupart, P., and Brubaker, M. (2019).
\newblock {Time2Vec}: Learning a vector representation of time.
\newblock {\em arXiv preprint arXiv:1907.05321}.

\bibitem[Kim et~al., 2023]{kim2023time}
Kim, J., Kang, H., and Kang, P. (2023).
\newblock Time-series anomaly detection with stacked transformer
  representations and 1d convolutional network.
\newblock {\em Engineering Applications of Artificial Intelligence},
  120:105964.

\bibitem[Li et~al., 2021]{li2021stacking}
Li, W., Hu, W., Chen, N., and Feng, C. (2021).
\newblock Stacking {VAE} with graph neural networks for effective and
  interpretable time series anomaly detection.
\newblock {\em arXiv preprint arXiv:2105.08397}.

\bibitem[Lin et~al., 2013]{lin2013network}
Lin, M., Chen, Q., and Yan, S. (2013).
\newblock Network in network.
\newblock {\em arXiv preprint arXiv:1312.4400}.

\bibitem[Liu et~al., 2021]{liu2021gated}
Liu, M., Ren, S., Ma, S., Jiao, J., Chen, Y., Wang, Z., and Song, W. (2021).
\newblock Gated transformer networks for multivariate time series
  classification.
\newblock {\em arXiv preprint arXiv:2103.14438}.

\bibitem[Liu et~al., 2022]{liu2022time}
Liu, S., Zhou, B., Ding, Q., Hooi, B., Zhang, Z., Shen, H., and Cheng, X.
  (2022).
\newblock Time series anomaly detection with adversarial reconstruction
  networks.
\newblock {\em IEEE Transactions on Knowledge and Data Engineering},
  35(4):4293--4306.

\bibitem[Malhotra et~al., 2016]{malhotra2016lstm}
Malhotra, P., Ramakrishnan, A., Anand, G., Vig, L., Agarwal, P., and Shroff, G.
  (2016).
\newblock Lstm-based encoder-decoder for multi-sensor anomaly detection.
\newblock {\em arXiv preprint arXiv:1607.00148}.

\bibitem[Malhotra et~al., 2015]{malhotra2015long}
Malhotra, P., Vig, L., Shroff, G., and Agarwal, P. (2015).
\newblock Long short term memory networks for anomaly detection in time series.
\newblock In {\em Proceedings}, volume~89, pages 89--94.

\bibitem[Mao et~al., 2019]{mao2019learning}
Mao, W., Liu, M., Salzmann, M., and Li, H. (2019).
\newblock Learning trajectory dependencies for human motion prediction.
\newblock In {\em Proceedings of the IEEE/CVF International Conference on
  Computer Vision}, pages 9489--9497.

\bibitem[Meng et~al., 2019]{meng2019spacecraft}
Meng, H., Zhang, Y., Li, Y., and Zhao, H. (2019).
\newblock Spacecraft anomaly detection via transformer reconstruction error.
\newblock In {\em International Conference on Aerospace System Science and
  Engineering}, pages 351--362. Springer.

\bibitem[Munir et~al., 2018]{munir2018deepant}
Munir, M., Siddiqui, S.~A., Dengel, A., and Ahmed, S. (2018).
\newblock Deepant: A deep learning approach for unsupervised anomaly detection
  in time series.
\newblock {\em Ieee Access}, 7:1991--2005.

\bibitem[Murphy, 2012]{murphy2012machine}
Murphy, K.~P. (2012).
\newblock {\em Machine learning: a probabilistic perspective}.
\newblock MIT press.

\bibitem[Pan et~al., 2021]{pan2021spatiotemporal}
Pan, C., Chen, S., and Ortega, A. (2021).
\newblock Spatio-temporal graph scattering transform.
\newblock In {\em International Conference on Learning Representations}.

\bibitem[Pang et~al., 2021]{pang2021deep}
Pang, G., Shen, C., Cao, L., and Hengel, A. V.~D. (2021).
\newblock Deep learning for anomaly detection: A review.
\newblock {\em ACM Computing Surveys (CSUR)}, 54(2):1--38.

\bibitem[Pedregosa et~al., 2011]{scikit-learn}
Pedregosa, F., Varoquaux, G., Gramfort, A., Michel, V., Thirion, B., Grisel,
  O., Blondel, M., Prettenhofer, P., Weiss, R., Dubourg, V., Vanderplas, J.,
  Passos, A., Cournapeau, D., Brucher, M., Perrot, M., and Duchesnay, E.
  (2011).
\newblock Scikit-learn: Machine learning in {P}ython.
\newblock {\em Journal of Machine Learning Research}, 12:2825--2830.

\bibitem[Su et~al., 2019]{su2019robust}
Su, Y., Zhao, Y., Niu, C., Liu, R., Sun, W., and Pei, D. (2019).
\newblock Robust anomaly detection for multivariate time series through
  stochastic recurrent neural network.
\newblock In {\em Proceedings of the 25th ACM SIGKDD international conference
  on knowledge discovery \& data mining}, pages 2828--2837.

\bibitem[Tran et~al., 2015]{tran2015learning}
Tran, D., Bourdev, L., Fergus, R., Torresani, L., and Paluri, M. (2015).
\newblock Learning spatiotemporal features with {3D} convolutional networks.
\newblock In {\em Proceedings of the IEEE international conference on computer
  vision}, pages 4489--4497.

\bibitem[Tuli et~al., 2022]{tuli2022tranad}
Tuli, S., Casale, G., and Jennings, N.~R. (2022).
\newblock Tranad: Deep transformer networks for anomaly detection in
  multivariate time series data.
\newblock {\em arXiv preprint arXiv:2201.07284}.

\bibitem[Vaswani et~al., 2017]{vaswani2017attention}
Vaswani, A., Shazeer, N., Parmar, N., Uszkoreit, J., Jones, L., Gomez, A.~N.,
  Kaiser, {\L}., and Polosukhin, I. (2017).
\newblock Attention is all you need.
\newblock In {\em Advances in neural information processing systems}, pages
  5998--6008.

\bibitem[Xu et~al., 2021]{xu2021anomaly}
Xu, J., Wu, H., Wang, J., and Long, M. (2021).
\newblock Anomaly transformer: Time series anomaly detection with association
  discrepancy.
\newblock {\em arXiv preprint arXiv:2110.02642}.

\bibitem[Yan et~al., 2021]{yan2021learning}
Yan, B., Peng, H., Fu, J., Wang, D., and Lu, H. (2021).
\newblock Learning spatio-temporal transformer for visual tracking.
\newblock {\em arXiv preprint arXiv:2103.17154}.

\bibitem[Yu et~al., 2020]{yu2020spatio}
Yu, C., Ma, X., Ren, J., Zhao, H., and Yi, S. (2020).
\newblock Spatio-temporal graph transformer networks for pedestrian trajectory
  prediction.
\newblock In {\em European Conference on Computer Vision}, pages 507--523.
  Springer.

\bibitem[Zhang et~al., 2020]{zhang2020tapnet}
Zhang, X., Gao, Y., Lin, J., and Lu, C.-T. (2020).
\newblock Tapnet: Multivariate time series classification with attentional
  prototypical network.
\newblock In {\em Proceedings of the AAAI Conference on Artificial
  Intelligence}, volume~34, pages 6845--6852.

\bibitem[Zhang et~al., 2023]{zhang2023stad}
Zhang, Z., Li, W., Ding, W., Zhang, L., Lu, Q., Hu, P., Gui, T., and Lu, S.
  (2023).
\newblock Stad-gan: Unsupervised anomaly detection on multivariate time series
  with self-training generative adversarial networks.
\newblock {\em ACM Transactions on Knowledge Discovery from Data}, 17(5):1--18.

\bibitem[Zhou et~al., 2021]{zhou2021informer}
Zhou, H., Zhang, S., Peng, J., Zhang, S., Li, J., Xiong, H., and Zhang, W.
  (2021).
\newblock Informer: Beyond efficient transformer for long sequence time-series
  forecasting.
\newblock In {\em Proceedings of AAAI}.

\bibitem[Zhou et~al., 2022]{zhou2022fedformer}
Zhou, T., Ma, Z., Wen, Q., Wang, X., Sun, L., and Jin, R. (2022).
\newblock {FEDformer}: Frequency enhanced decomposed transformer for long-term
  series forecasting.
\newblock {\em arXiv preprint arXiv:2201.12740}.

\end{thebibliography}

\newpage

\appendix


\section{Elevator use-case} \label{app:elevator_diagram}

\begin{figure}
    \centering
    \includegraphics[width = 0.7\textwidth]{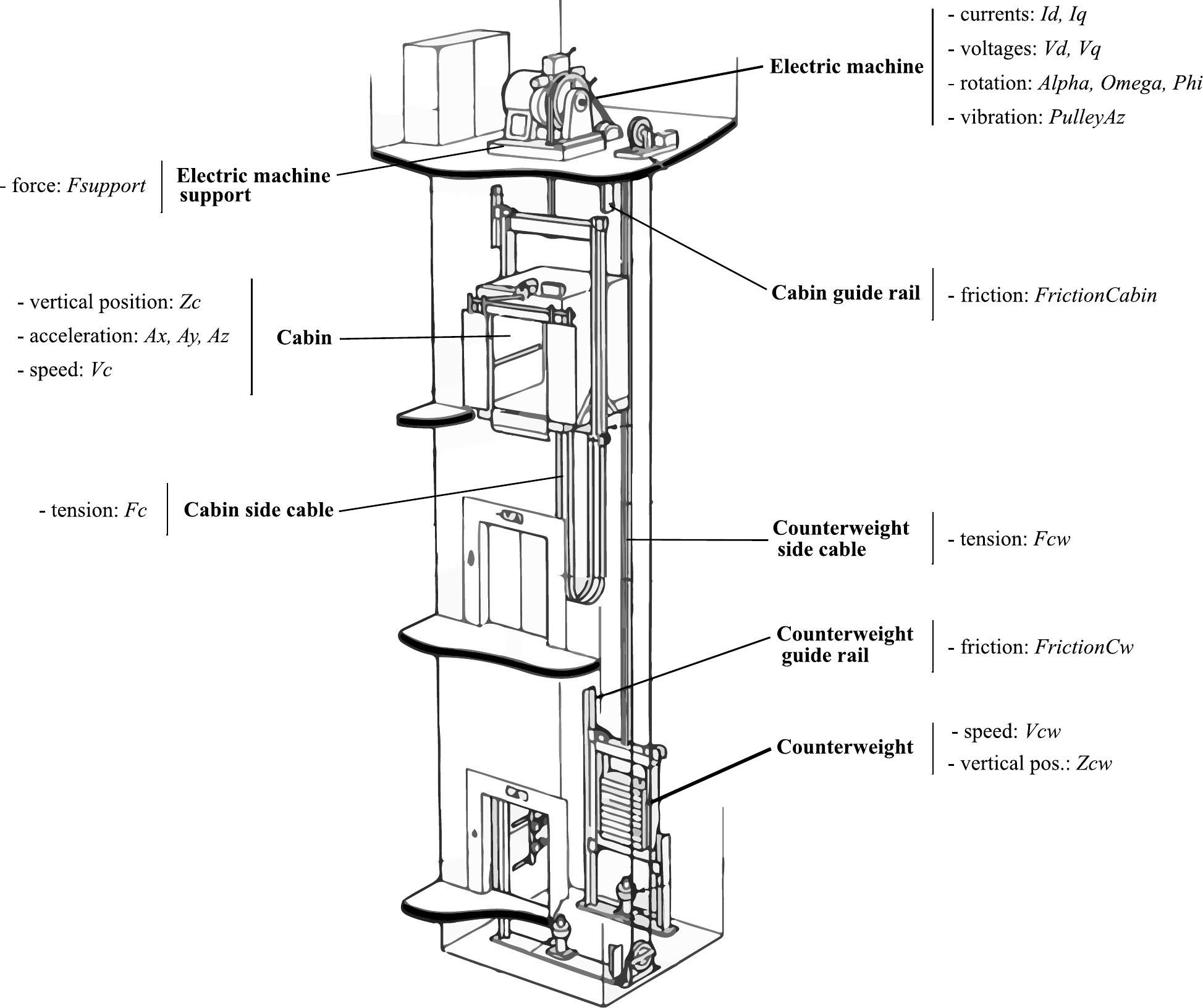}
    \caption{Diagram of the elevator.}
    \label{fig:elevator}
\end{figure}

The industrial use case of this work deals with an elevator monitored by 20 sensors. Figure \ref{fig:elevator}  shows a diagram of the elevator with its respective sensors. The data used in this work is obtained from a physical model designed by a domain expert that mimics the real behavior of the elevator. This simulator is highly configurable, having different parameters to determine the power of the electric machine, the load of the cabin, the alignment of the guidance system, or the tension of the cables, among others. The physical model simulates up and down journeys under all possible conditions by modifying these input parameters. Specific values of these parameters allow generating fault conditions during the journey, such as reduced lubrication, misalignment or peaks in the guiding system, or de-magnetization or loss of inductance in the electric machine. This paper studies the most common issue in this case study: the failures produced in the guiding system. In order to generate these anomalies, as pointed out before, three effects have been introduced to the model related to misalignment, lubrication reduction, and localized bumps in the guiding system. The first two effects increase friction between the cabin, the counterweight, and the rails. The localized bumps result in local impacts on the cabin, leading to spike-like malformations. These effects cover the main fault conditions of the guiding systems. All failures can be observed in several of the described sensors, but they may not be reflected in all of them.

\section{1D Multi-Head CNN}\label{app:mh1dcnn}



\subsection{Architecture and hyperparameters}
\begin{minipage}{\textwidth}
  \begin{minipage}[!ht]{0.49\textwidth}
    \centering
    \includegraphics[width=\textwidth]{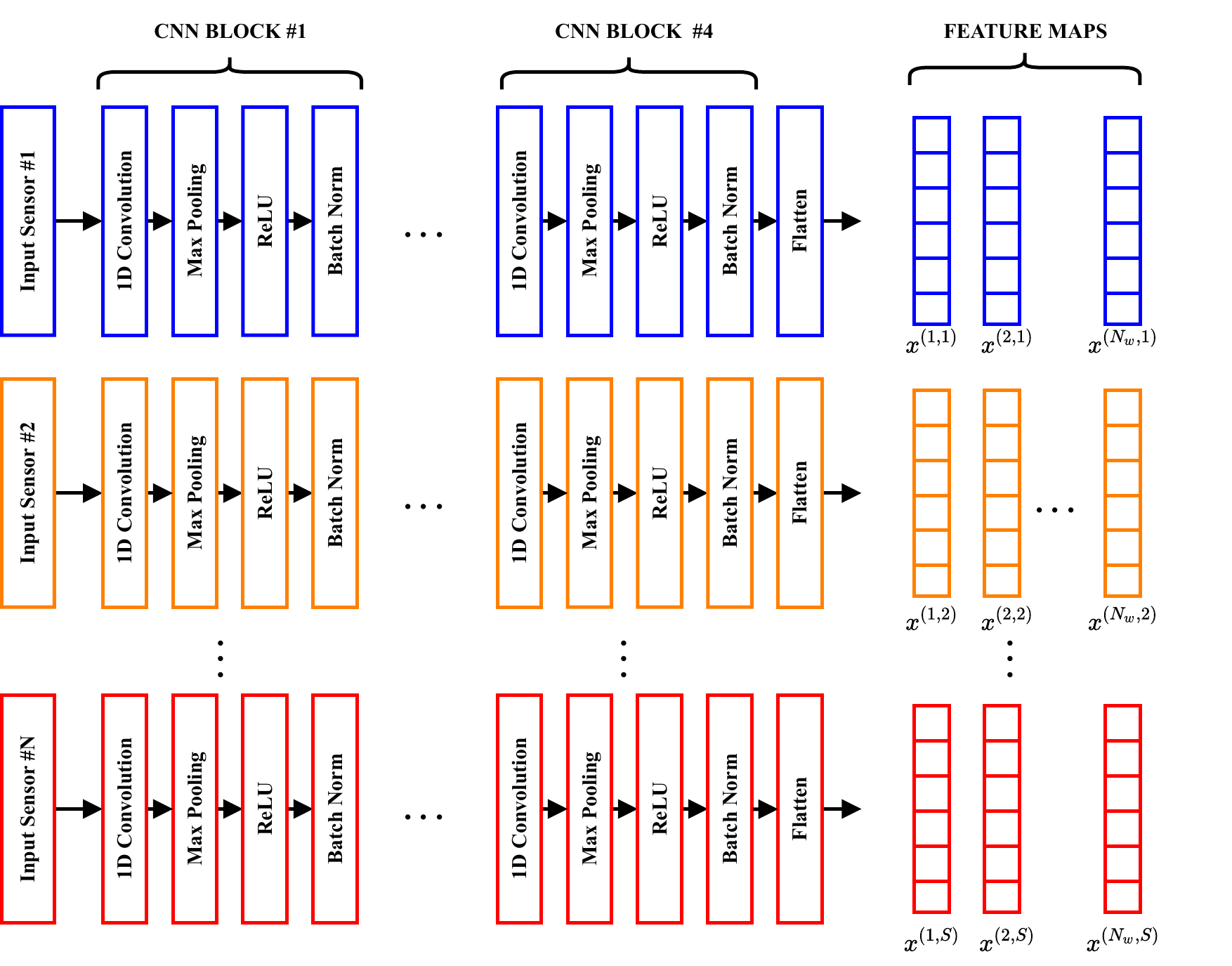}
    \captionof{figure}{1D Multi-Head CNN used for feature extraction.} 
    \label{fig:1d_multihead}
  \end{minipage}
  \begin{minipage}[!ht]{0.49\textwidth}
  
        \captionof{table}{Hyperparameters for Multi-Head 1D CNN.}
        \label{tab:mh1dcnn}
        \centering
        \resizebox{\textwidth}{!}{%
        \begin{tabular}{l|cccc}
        \toprule[1.5pt]
        \multicolumn{1}{c|}{\multirow{2}{*}{\textbf{Parameter}}} & \multicolumn{4}{c}{\textbf{Datasets}}                           \\ \cline{2-5} 
        \multicolumn{1}{c|}{}                                    & \textbf{Elevator} & \textbf{SMD} & \textbf{MSL} & \textbf{SMAP} \\ \midrule
        \textit{Number of sensors (S)}          & 20  & 38  & 55  & 25  \\
        \textit{Window length ($w_l$)}          & 100 & 50  & 50  & 50  \\
        \textit{Time-segments ($N_w$)}             & 80  & 10  & 10  & 10  \\
        \textit{Number of convolutional blocks} & 4   & 4   & 4   & 4   \\
        \textit{Kernel size}                    & 5   & 5   & 5   & 5   \\
        \textit{Embedding dim (d)}              & 240 & 120 & 120 & 120 \\ \textit{Pooling size}              & 2 & 2 & 2 & 2 \\
        \textit{Pooling stride}              & 2 & 2 & 2 & 2 \\\bottomrule[1.5pt]
        \end{tabular}%
        }                
    \end{minipage}
    \vspace{5mm}
  \end{minipage}
  
 Figure \ref{fig:1d_multihead} shows the architecture of the 1D Multi-Head CNN used in DFStrans and Strans.  The only difference concerning the blocks proposed by Cañizo et al.\cite{Canizo2019} is the introduction of Max Pooling layers. As shown in Figure \ref{fig:1d_multihead}, each convolution applies a Max Pooling layer, followed by a ReLU activation function and a Batch Normalization (BN) \cite{ioffe2015batch} layer. The Max Pooling layer helps to significantly reduce the dimensionality of the extracted features by statistically summarizing the output of the convolutional layers at a given time point based on their neighbors \cite{munir2018deepant}.

 Table \ref{tab:mh1dcnn} shows the details of the data used, such as the number of sensors, the length of the windows or the number of time-segments and also shows the hyperparameters used in each dataset for the 1D Multi-Head CNN. These are chosen based on the exhaustive analysis made by \cite{Canizo2019} in their paper.
 
\subsection{Comparison with linear projection}

Next we want to measure the effect that the Multi-Head 1D CNN has on the performance of DFStrans in comparison to whether the embeddings that are fed to the spatio-temporal dependency discovery network are achieved by linear projections, as done in \cite{aksan2020spatio}, and non-linear projection, as done in \cite{liu2021gated}.

As stated, the MH 1D CNN maps each raw time-series $\tau$-segment to a representation matrix: $U_i^{(\tau)} \in \mathbb{R}^{S \times w_L} \to \sfX_i^{(\tau)} \in \mathbb{R}^{S\times M}$, where $w_l$ denotes the window length and $M$ is the dimensionality of the embedding. These embeddings are obtained by means of different convolutional blocks in our case. But in \cite{aksan2020spatio} for example, these embeddings are linear projections of the raw data, i.e.

\begin{equation}
    \sfX_i^{(\tau)} \triangleq  U_i^{(\tau)} \sfW^{(\tau)} ,
\end{equation}

or non-linear projections, as in \cite{liu2021gated}, i.e.

\begin{equation}
    \sfX_i^{(\tau)} \triangleq \tanh(U_i^{(\tau)}\sfW^{(\tau)} ),
\end{equation}

where $\sfW^{(\tau)} \in \mathbb{R}^{w_l\times M}$ is a fully learneable matrix.

\begin{table}[!ht]
\centering
\caption{Comparison between DFStrans using Multi-Head 1DCNN and DFStrans with linearly and non-linearly projected embeddings.}
\label{tab:linear_comp}
\resizebox{0.7\textwidth}{!}{%
\begin{tabular}{c|ccc}
\toprule[1.5pt]
\multicolumn{1}{c|}{\textbf{Feature extraction}}     & \textbf{Precision} & \multicolumn{1}{c}{\textbf{Recall}} & \multicolumn{1}{c}{\textbf{F1}} \\ \midrule
Multi-Head 1D CNN                      &\textbf{0.989 }$\pm$ \textbf{0.016}        & \textbf{ 0.917}  $\pm$   \textbf{0.022}                      & \textbf{0.952 } $\pm$  \textbf{0.005}                   \\
Linear projection &   0.986 $\pm$ 0.002             &          0.668 $\pm$    0.045                      &    0.778  $\pm$ 0.041                              \\
Non-linear projection &     0.980 $\pm$ 0.002               &         0.85 $\pm$ 0.043                            &        0.912 $\pm$ 0.030                         \\\bottomrule[1.5pt]
\end{tabular}%
}

\end{table}

Table \ref{tab:linear_comp} shows the results using different feature extraction strategies to embed the raw time-series. This experiment has been performed for the elevator use case. Looking at the results, we see that embedding extraction plays an essential role in detecting anomalies. Using Multi-Head 1D CNN as a feature extractor, we outperform the other strategies in terms of Recall and F1. Although the Precision is still high with the other methods, it loses detection capability, and the variability is higher too. Among the other two strategies, non-linear projections performed better than linear projections.


\section{Visualization of positional encodings} \label{app:dft}

\begin{figure*}[!ht]
     \centering
     \begin{subfigure}[b]{0.7\textwidth}
         \centering
         \includegraphics[width=\textwidth]{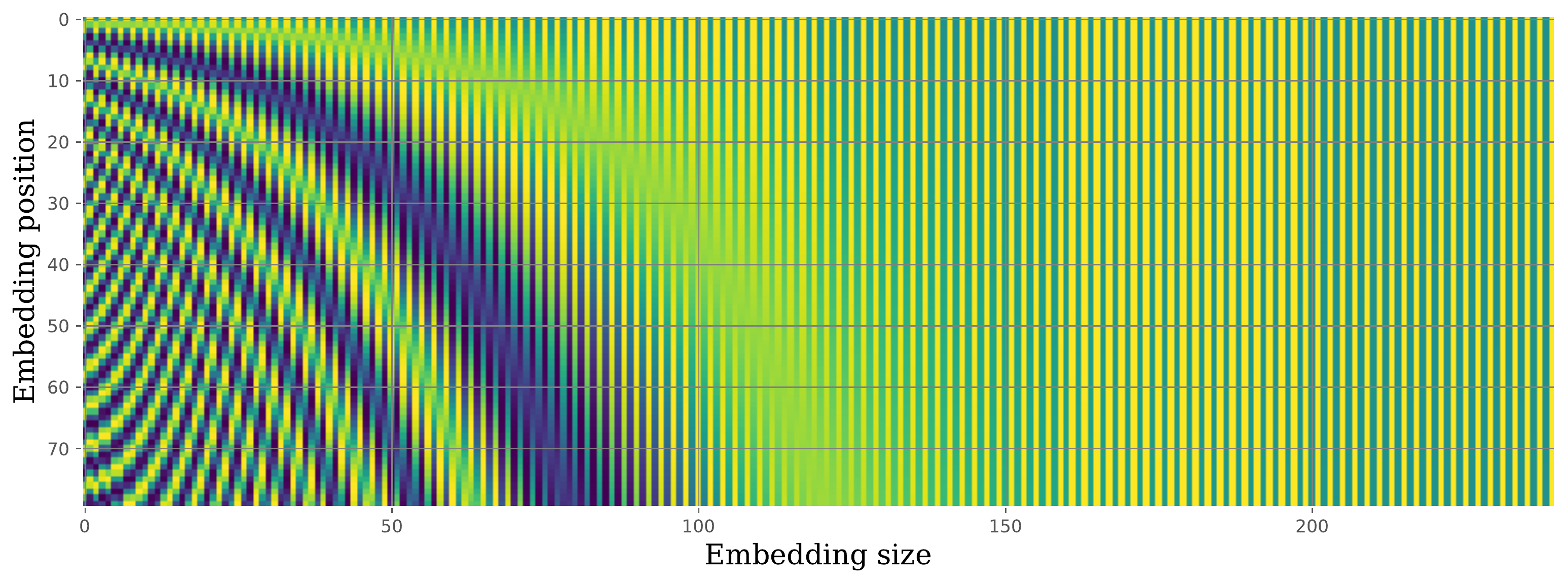}
         \phantomsubcaption
         \label{fig:vanilla_pe}
     \end{subfigure}\hfill
     \begin{subfigure}[b]{0.7\textwidth}
         \centering
         \includegraphics[width=\textwidth]{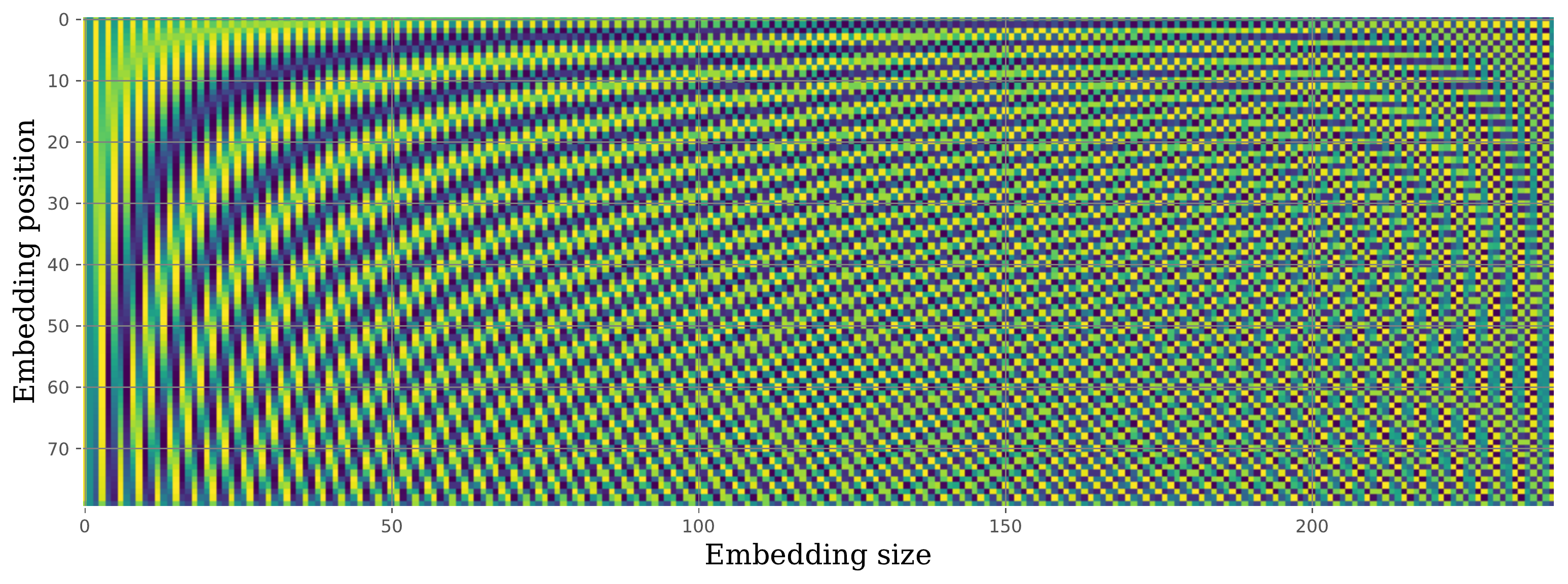}
        \phantomsubcaption
        \label{fig:dft_pe}
     \end{subfigure}

    \caption{The figure on \textbf{top} shows a visualization of vanilla positional encoding and the figure of the \textbf{bottom} shows a visualization of \textbf{faithful-Encoding}. Both for $N_w = 80$ and $d=240$.}
    \label{fig:pe_comparisons}
    
\end{figure*}

In Figure \ref{fig:pe_comparisons} we have visualized, on the one hand, the vanilla positional encoding, and on the other hand, the faithful-Encoding that we propose. In this example, we have used $N_w=80$ time-segments and $d=240$ features.

\section{Details of spatio-temporal dependency structure and classification head}\label{app:ST-dependency_model}
Table \ref{tab:trans_hyp} shows the hyperparameters used in the proposed network. Although the experiments can be generalized to more heads and more layers in the spatio-temporal dependency structure, we focus on a single layer and a single attention head to facilitate the interpretation of the network for anomaly diagnosis. The number of units for the spatial and temporal attention feedforward layer is 2048. The activations we use in the network are generally ReLUs, except for the last classification layer, which is a sigmoid. Moreover, the dropout rate is 0.1 in the whole network. 

\begin{table}[!ht]
\centering
\caption{The hyperparameters selected for the spatio-temporal dependency structure. Same selection for every dataset.}
\label{tab:trans_hyp}
\resizebox{0.6\textwidth}{!}{%

\begin{tabular}{ll|c}
\toprule[1.5pt]
\multicolumn{2}{c|}{\textbf{Hyperparameters}}                 & \textbf{Value} \\ \midrule
\multicolumn{1}{c|}{\multirow{5}{*}{\textbf{ST dependency structure}}} & \textit{Dim feedforward} & 2048 \\
\multicolumn{1}{c|}{} & \textit{Dropout rate}                & 0.1            \\
\multicolumn{1}{c|}{} & \textit{Number of attention heads}   & 10              \\
\multicolumn{1}{c|}{} & \textit{Number of layers}            & 1              \\ 
\multicolumn{1}{c|}{} & \textit{Activations}            & ReLU              \\ \midrule
\multicolumn{1}{l|}{\multirow{5}{*}{\textbf{Classification head}}} & \textit{Dim feedforward 1}                      & 512  \\
\multicolumn{1}{l|}{} & \textit{Activation in feedforward 1} & ReLU           \\
\multicolumn{1}{l|}{} & \textit{Dropout rate}                & 0.1            \\
\multicolumn{1}{l|}{} & \textit{Dim feedforward 2}           & 1              \\
\multicolumn{1}{l|}{} & \textit{Activation in feedforward 2} & Sigmoid        \\ \bottomrule[1.5pt]

\end{tabular}%
}

\end{table}

\paragraph{\textbf{Justification of the transation probability equations}} These are the equations that we are going to justify:
\begin{align}
p(\tau,s \mid \tau',s') \approx p(\tau\mid \tau',s)p(s\mid s',\tau), \label{eq:p1}
\end{align}
 and
\begin{align}
    \ln p(\tau\mid \tau',s) &= \mathrm{const.} + \frac{1}{\sqrt{M}} (\hat{\bmx}^{(\tau,s)})^\top \hat{\sfH}\hat{\bmx}^{(\tau',s)}, \label{eq:ln1}\\
    \ln p(s\mid s',\tau) &= \mathrm{const.} + \frac{1}{\sqrt{M}}(\overline{\bmx}^{(\tau,s)})^\top \overline{\sfH}\overline{\bmx}^{(\tau,s')}\label{eq:ln2},
\end{align}

The key motivation of Eq. \ref{eq:p1} is how to take account of the dependency between spatial and temporal coordinates. The most naive approach is to treat them as independent like
\begin{equation}
p\left(\tau, s \mid \tau^{\prime}, s^{\prime}\right) \approx p\left(\tau \mid \tau^{\prime}, s^{\prime}\right) p\left(s \mid \tau^{\prime}, s^{\prime}\right) \quad \textbf{(naive)}
\end{equation}

However, this is not the best model because we know that different sensors ($s,s'$) have different time correlations, and spatial dependency can vary at different times ($\tau,\tau'$). Thus, Eq.\ref{eq:p1} moves one step ahead of this naive model and can be viewed as a tractable but still tractable approximation of the full model. Instead of naively assuming independence, we used
 \begin{equation}
p\left(\tau, s \mid \tau^{\prime}, s^{\prime}\right) \approx p\left(\tau \mid \tau^{\prime}, s\right) p\left(s \mid \tau, s^{\prime}\right) \quad \textbf{(ours)}
\end{equation}
so that the dependency between $s$ and $\tau$ is captured to some extent. Moreover, the Eqs. \ref{eq:ln1} and \ref{eq:ln2} have two relatively clear justifications. 
\begin{enumerate}
    \item It leads to the well-known query-key formalism of the transformer. 
    \item  Eqs.\ref{eq:ln1} and \ref{eq:ln2} amount to approximating the distribution up to the second order moments. This is indeed a widely-used technique. For instance, you can think of it as a Laplace approximation (see \cite{murphy2012machine}) Section 8.4.1), where the idea is essentially "use the 2nd order Taylor expansion of $\ln p$".
\end{enumerate}
\section{Details of benchmark algorithms}
\paragraph{Used code.} Here are the links to the codes for the benchmark datasets: 
\begin{itemize}
    \item \textbf{InceptionTime:} \url{https://github.com/TheMrGhostman/InceptionTime-Pytorch}
    \item \textbf{TapNet:} \url{https://github.com/xuczhang/tapnet}
    \item \textbf{MLSTM-FCN:} \url{https://github.com/metra4ok/MLSTM-FCN-Pytorch}
    \item \textbf{Multi-Head 1D CNN - LSTM: } No open source code has been found for this algorithm.
\end{itemize}

\paragraph{\textbf{Hyperparameters}} Table \ref{tab:hyp_benchmark} shows the hyperparameters selected in each benchmark algorithm for each dataset. In most cases the hyperparameters selected are the default parameters that the algorithms provide, but there have been some that we have considered changing. For example, in the case of TapNet, the \textit{dilatation} and \textit{rp params} parameters depend on the input dimension and the length of the series, so it varies for each case. The \textit{rp param} consist of two parameters: the first one corresponds to the number of permutations, and the second one to the sub-dimension of each permutation. Following the official implementation, the number of permutations is set to three and the sub-dimension is computed as
$$
\lfloor S/1.5 \rfloor,
$$
where $S$ denotes the number of sensors. Moreover, the \textit{dilatation} parameter is the dilatation used in the first dilated convolution, and is computed as
$$
\lfloor T/64 \rfloor,
$$
where $T$ denotes the length of the time series. On the other hand, another parameter that we have had to tune has been the kernel sizes in InceptionTime. Here we have differentiated the elevator use case and the benchmark datasets because the difference in the length of the series is considerable. On the one hand, in the elevator use case, the length of the series is 8000 points, so we have considered large kernels of size 49, 99, and 199 in this case. On the other hand, in the benchmark datasets, as the series are of 500 points, we have considered kernels of size 9, 19, and 39.

\begin{landscape}

\begin{table}[]
\centering
\caption{Hyperparameters selected for benchmark algorithms.}
\label{tab:hyp_benchmark}
\resizebox{1.5\textwidth}{!}{%
\begin{tabular}{l|l|cccc}
\toprule[1.5pt]
\multicolumn{1}{c|}{\multirow{2}{*}{\textbf{Algorithm}}} &
  \multicolumn{1}{c|}{\multirow{2}{*}{\textbf{Hyperparameter}}} &
  \multicolumn{4}{c}{\textbf{Dataset}} \\ \cline{3-6} 
\multicolumn{1}{c|}{} &
  \multicolumn{1}{c|}{} &
  \multicolumn{1}{c|}{\textbf{Elevator}} &
  \multicolumn{1}{c|}{\textbf{SMD}} &
  \multicolumn{1}{c|}{\textbf{MSL}} &
  \textbf{SMAP} \\ \midrule
\textbf{MLSTM-FCN} &
  \textit{\begin{tabular}[c]{@{}l@{}}Conv 1 (filters, kernel size, stride)\\ Conv 2 (filters, kernel size, stride)\\ Conv 3 (filters, kernel size, stride)\\ Conv dropout rate\\ Lstm layers\\ Lstm units \\ Lstm dropout rate\\ Dim feedforward\end{tabular}} &
  \multicolumn{1}{c|}{\begin{tabular}[c]{@{}c@{}}(128,8,1)\\ (256,5,1)\\ (128,3,1)\\ 0.3\\ 1\\ 128\\ 0.8\\ 128\end{tabular}} &
  \multicolumn{1}{c|}{\begin{tabular}[c]{@{}c@{}}(128,8,1)\\ (256,5,1)\\ (128,3,1)\\ 0.3\\ 1\\ 128\\ 0.8\\ 128\end{tabular}} &
  \multicolumn{1}{c|}{\begin{tabular}[c]{@{}c@{}}(128,8,1)\\ (256,5,1)\\ (128,3,1)\\ 0.3\\ 1\\ 128\\ 0.8\\ 128\end{tabular}} &
  \begin{tabular}[c]{@{}c@{}}(128,8,1)\\ (256,5,1)\\ (128,3,1)\\ 0.3\\ 1\\ 128\\ 0.8\\ 128\end{tabular} \\ \midrule
\textbf{TapNet} &
  \textit{\begin{tabular}[c]{@{}l@{}}Conv 1 (filters, kernel size, stride)\\ Conv 2 (filters, kernel size, stride)\\ Conv 3 (filters, kernel size, stride)\\ Lstm units \\ Dilatation  \\ Rp params\end{tabular}} &
  \multicolumn{1}{c|}{\begin{tabular}[c]{@{}c@{}}(256,8,1)\\ (256,5,1)\\ (128,3,1)\\ 128\\ 125\\ (3,13)\end{tabular}} &
  \multicolumn{1}{c|}{\begin{tabular}[c]{@{}c@{}}(256,8,1)\\ (256,5,1)\\ (128,3,1)\\ 128\\ 7\\ (3,25)\end{tabular}} &
  \multicolumn{1}{c|}{\begin{tabular}[c]{@{}c@{}}(256,8,1)\\ (256,5,1)\\ (128,3,1)\\ 128\\ 7\\ (3,36)\end{tabular}} &
  \begin{tabular}[c]{@{}c@{}}(256,8,1)\\ (256,5,1)\\ (128,3,1)\\ 128\\ 7\\ (3,16)\end{tabular} \\ \midrule
\textbf{InceptionTime} &
  \textit{\begin{tabular}[c]{@{}l@{}}Inception blocks\\ Block 1 (in channels, filters, kernel sizes,\\ bottleneck channels, use residual, activation)\\ Block 2 (in channels, filters, kernel sizes,\\ bottleneck channels, use residual, activation)\\ Block 3 (in channels, filters, kernel sizes,\\ bottleneck channels, use residual, activation)\\ Maxpool (kernel size, stride)\end{tabular}} &
  \multicolumn{1}{c|}{\begin{tabular}[c]{@{}c@{}}3\\ (20,32,[49,99,199],32,True,ReLU)\\ \\ (128,32,[49,99,199],32,True,ReLU)\\ \\ (128,32,[49,99,199],32,True,ReLU)\\ \\ (3,1)\end{tabular}} &
  \multicolumn{1}{c|}{\begin{tabular}[c]{@{}c@{}}3\\ (38,32,[9,19,39],32,True,ReLU)\\ \\ (128,32,[9,19,39],32,True,ReLU)\\ \\ (128,32,[9,19,199],32,True.ReLU)\\ \\ (3,1)\end{tabular}} &
  \multicolumn{1}{c|}{\begin{tabular}[c]{@{}c@{}}3\\ (55,32,[9,19,39],32,True,ReLU)\\ \\ (128,32,[9,19,39],32,True,ReLU)\\ \\ (128,32,[9,19,39],32,True,ReLU)\\ \\ (3,1)\end{tabular}} &
  \begin{tabular}[c]{@{}c@{}}3\\ (25,32,[9,19,39],32,True,ReLU)\\ \\ (128,32,[9,19,39],32,True,ReLU)\\ \\ (128,32,[9,19,39],32,True,ReLU)\\ \\ (3,1)\end{tabular} \\ \midrule
\textbf{MH 1DCNN} &
  \textit{\begin{tabular}[c]{@{}l@{}}Convolutional blocks\\ Conv  (filters, kernel size, stride)\\ Maxpool (kernel size, stride) \\ Dropout rate\\ Lstm units\end{tabular}} &
  \multicolumn{1}{c|}{\begin{tabular}[c]{@{}c@{}}3\\ (20,5,1)\\ (2,2)\\ 0.1\\ 128\end{tabular}} &
  \multicolumn{1}{c|}{\begin{tabular}[c]{@{}c@{}}3\\ (20,5,1)\\ (2,2)\\ 0.1\\ 128\end{tabular}} &
  \multicolumn{1}{c|}{\begin{tabular}[c]{@{}c@{}}3\\ (20,5,1)\\ (2.2)\\ 0.1\\ 128º\end{tabular}} &
  \begin{tabular}[c]{@{}c@{}}3\\ (20,5,1)\\ (2,2)\\ 0.1\\ 128\end{tabular} \\ \bottomrule[1.5pt]
\end{tabular}%
}

\end{table}
\end{landscape}

\section{Discussion on number of parameters and training time}

\subsection{Number of parameters and training time in the conducted experiments}
Table \ref{tab:timeparams} shows each algorithm's learnable parameters and the time it takes to train for each epoch. As we can see, in the elevator use-case, MH1DCNN-LSTM is the model that has the most parameters, followed by DFStrans, which is the model with most parameters in the other use cases. Genarally, training DFSTrans takes longer than the other algorithms except in the elevator use-case, where MLSTM-FCN is the slowest algorithm.t. The algorithm that has fewer parameters to train is MLSTM-FCN. The fastest algorithms are TapNet and InceptionTime. 

\begin{table}[!ht]
\centering
\caption{Training time and number of parameters.}
\label{tab:timeparams}
\resizebox{\textwidth}{!}{%
\begin{tabular}{l|l|c|c}
\toprule[1.5pt]
\textbf{Algorithm} &
  \textbf{Dataset} &
  \textbf{Number of parameters} &
  \textbf{Training time (s/epochs)} \\ \midrule
\textbf{DFStrans} &
  \begin{tabular}[c]{@{}l@{}}Elevator\\ SMD\\ SMAP\\ MSL\end{tabular} &
  \begin{tabular}[c]{@{}c@{}}3878033\\ 5840153\\ 2225553\\ 4197153\end{tabular} &
  \begin{tabular}[c]{@{}c@{}}102.14\\ 3.36\\ 19.24\\ 8.11\end{tabular} \\ \midrule
\textbf{InceptionTime} &
  \begin{tabular}[c]{@{}l@{}}Elevator\\ SMD\\ SMAP\\ MSL\end{tabular} &
  \begin{tabular}[c]{@{}c@{}}3303681\\ 726657\\ 724161\\ 729921\end{tabular} &
  \begin{tabular}[c]{@{}c@{}}131.29\\ 2.10\\ 10.48\\ 2.54\end{tabular} \\ \midrule
\textbf{TapNet} &
  \begin{tabular}[c]{@{}l@{}}Elevator\\ SMD\\ SMAP\\ MSL\end{tabular} &
  \begin{tabular}[c]{@{}c@{}}5155962\\ 1389690\\ 1334394\\ 1457274\end{tabular} &
  \begin{tabular}[c]{@{}c@{}}96.83\\ 2.62\\ 8.23\\ 2.32\end{tabular} \\ \midrule
\textbf{MLSTM-FCN} &
  \begin{tabular}[c]{@{}l@{}}Elevator\\ SMD\\ SMAP\\ MSL\end{tabular} &
  \begin{tabular}[c]{@{}c@{}}371457\\ 399105\\ 379137\\ 425217\end{tabular} &
  \begin{tabular}[c]{@{}c@{}}203.21\\ 2.49\\ 9.81\\ 2.66\end{tabular} \\ \midrule
\textbf{MH1DCNN-LSTM} &
  \begin{tabular}[c]{@{}l@{}}Elevator\\ SMD\\ SMAP\\ MSL\end{tabular} &
  \begin{tabular}[c]{@{}c@{}}20764385\\ 2696145\\ 1841785\\ 2768905\end{tabular} &
  \begin{tabular}[c]{@{}c@{}}73.48\\ 3.51\\ 25.32\\ 7.24\end{tabular} \\ \bottomrule[1.5pt]
\end{tabular}%
}

\end{table}

\subsection{Effect of data size on the number of parameters}

\begin{figure}[!ht]
     \centering
     \includegraphics[width=\textwidth]{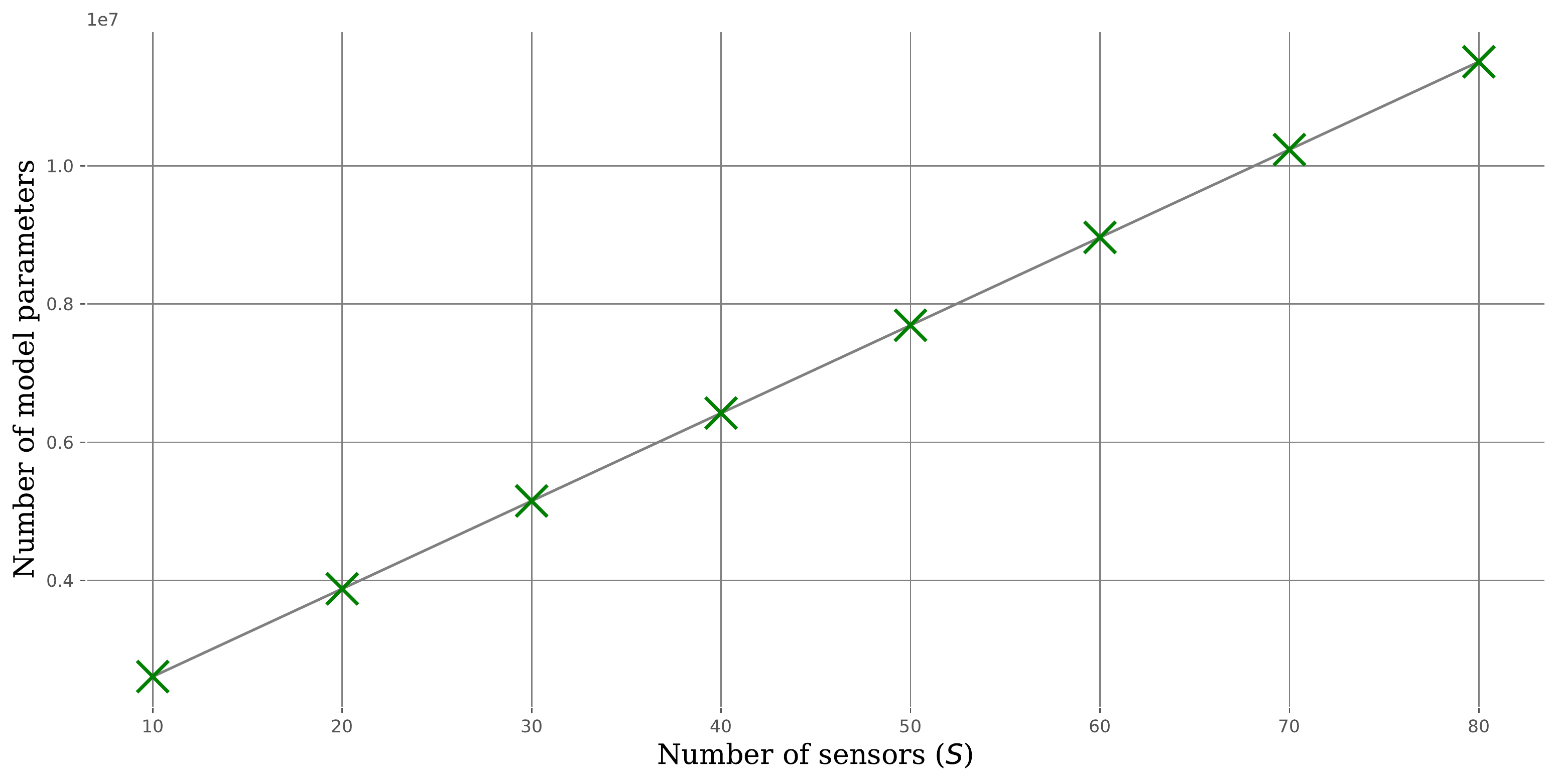}
    \caption{This figure shows the linear relationship between model parameters and $S$.}
    \label{fig:n_param_s}
\end{figure}

In this section, we study how the model parameters grow as the spatio-temporal dimensions of the multi-sensor systems grow. As the learneable parameters of the model does not depend on number of the time-segments, we have studied how the model's number of parameters changes with different number of sensors. Table \ref{tab:n_params_s} show the number of DFStrans parameters for different numbers sensors (being $N_w = 20$). These results have been plotted in Figure \ref{fig:n_param_s}, where it can be seen that the number of model parameters grows linearly with the number of sensors.

\begin{table}[ht!]
\centering
\resizebox{\textwidth}{!}{%
\begin{tabular}{c|cccccccc}
\toprule[1.5pt]
\textbf{Number of sensors, $N_w =20$} & \textbf{10}       & \textbf{20}       & \textbf{30}       & \textbf{40}       & \textbf{50}       & \textbf{60}       & \textbf{70}       & \textbf{80}       \\ \midrule
\textbf{Number of parameters    }    & 2606433 & 3878033 & 5149633 & 6421233 & 7692833 & 8964433 & 10236033 & 11507633 \\ \bottomrule[1.5pt]
\end{tabular}%
}
\caption{Number of model parameters for different number of sensors.}
\label{tab:n_params_s}
\end{table}

\section{Experimental framework}
The experiments are run on an Nvidia-Docker container that uses ubuntu 18.04. The models were implemented using on Pytorch. For training the models an NVIDIA TITAN V GPU has been used, with a memory of 12 GB, in an Intel i7-6850K 3.6Ghz machine with 32 GB of DDR4 RAM.

\section{Limitations and societal impacts}

Our algorithm is a supervised ST dependency discovery model designed to detect and primarily diagnose anomalies in multi-sensor data. The main limitation of this architecture is that labeled data is needed for training the model. Regarding societal impacts, DFStrans improves the understanding of the model with the diagnostics and may help to trust in the model's decisions. This can be essential to implementing AI in industrial environments where black boxes are not used right now.

\end{document}